\pdfoutput=1
\documentclass[11pt]{article}

\usepackage{emnlp2021}

\usepackage{times}
\usepackage{latexsym}
\usepackage[T1]{fontenc}
\usepackage[utf8]{inputenc}
\usepackage{microtype}

\usepackage{enumitem}
\setlist[itemize]{itemsep=2pt}
\setlist[enumerate]{itemsep=2pt}
\usepackage{todonotes}  
\usepackage{nicefrac}
\usepackage{amsmath}
\usepackage{amssymb}
\usepackage{amsfonts}
\usepackage{booktabs}
\usepackage{multicol}
\usepackage{blindtext}
\usepackage{floatrow}
\usepackage{caption}
\usepackage{subfig}
\usepackage{pifont}  
\usepackage{comment}
\usepackage{threeparttable}
\usepackage[hang,flushmargin]{footmisc}
\usepackage{cuted}

\newfloatcommand{capbtabbox}{table}[][\FBwidth]
\DeclareCaptionLabelFormat{andfigure}{#1~#2  \&  \figurename~\thefigure}

\renewcommand{\v}[1]{\ensuremath{\mathbf{#1}}}  
\newcommand{\m}[1]{\ensuremath{\mathbf{#1}}}  
\newcommand{\N}{\ensuremath{\mathbb{N}}}
\newcommand{\Z}{\ensuremath{\mathbb{Z}}}

\newcommand{\R}{\ensuremath{\mathbb{R}}}

\newcommand{\round}[1]{\ensuremath{\left\lfloor{#1}\right\rceil}}
\newcommand{\wh}[1]{\widehat{#1}}

\newcommand{\p}[1]{\left(#1\right)}
\renewcommand{\b}[1]{\left\lbrack#1\right\rbrack}

\renewcommand{\u}{\textbf}
\DeclareMathOperator*{\argsort}{arg\,sort}

\title{Understanding and Overcoming the Challenges of \\Efficient Transformer Quantization}

\author{Yelysei Bondarenko, Markus Nagel, Tijmen Blankevoort \\
    Qualcomm AI Research\thanks{\scriptsize \,\,\,\,Qualcomm AI Research is an initiative of Qualcomm Technologies, Inc.} \\
    \texttt{\{ybond, markusn, tijmen\}@qti.qualcomm.com} \\}

\begin{document}

    \newif\ifappendix
    \newif\ifcontent
    
    \appendixtrue  

\maketitle  

    
\begin{abstract}
%
Transformer-based architectures have become the de-facto standard models for a wide range of Natural Language Processing tasks.
However, their memory footprint and high latency are prohibitive for efficient deployment and inference on resource-limited devices.
In this work, we explore quantization for transformers.
We show that transformers have unique quantization challenges -- namely, high dynamic activation ranges that are difficult to represent with a low bit fixed-point format.
%
We establish that these activations contain structured outliers in the residual connections that encourage specific attention patterns, such as attending to the special separator token.
%
%
To combat these challenges, we present three solutions based on post-training quantization and quantization-aware training, each with a different set of compromises for accuracy, model size, and ease of use.
In particular, we introduce a novel quantization scheme -- per-embedding-group quantization.
%
%
We demonstrate the effectiveness of our methods on the GLUE benchmark using BERT, establishing state-of-the-art results for post-training quantization.
%
Finally, we show that transformer weights and embeddings can be quantized to ultra-low bit-widths, leading to significant memory savings with a minimum accuracy loss.
Our source code is available at~\url{https://github.com/qualcomm-ai-research/transformer-quantization}.
\end{abstract}

\section{Introduction}
\label{sec:intro}


Recently, transformer architectures have shown remarkable improvement in many Natural Language Processing (NLP) tasks and beyond.
Based on the original Transformer~\citep{vaswani2017attention}, language models pre-trained from large corpora of unlabeled text, such as BERT~\citep{devlin-etal-2019-bert}, RoBERTa~\citep{liu2019roberta}, XLNet~\citep{yang2019xlnet}, Transformer-XL~\citep{dai-etal-2019-transformer}, GPT family~\citep{radford2018improving,radford2019language,brown2020language}, have become an indispensable building block in modern NLP pipelines.
They are also increasingly adopted in other areas, including vision~\citep{carion2020end,dosovitskiy2020image,chen2020generative,girdhar2019video} and audio~\citep{dong2018speech,child2019generating}.

Despite cutting edge results in many applications, pre-trained transformer-based models are extremely large, sometimes exceeding billions of parameters.
Hence, efficient deployment of these models on resource-constrained embedded systems, and even sometimes in data centers, has become an important problem due to high latency and prohibitively large memory footprint and energy consumption.

One effective method to tackle this problem is neural network quantization. Quantization
 reduces memory consumption by using low-bit precision for weight and activation tensors.
Is also reduces inference time, and improves energy efficiency by employing low-bit fixed-point arithmetic instead of floating-point arithmetic~\citep{horowitz}.

Quantization, however, is not free. 
It introduces additional noise in the network that can lead to a drop in the model's performance.
While prior work has demonstrated the feasibility of integer-only inference for computer vision models~\citep{lin2016fixed,jacob2018quantization,krishnamoorthi2018quantizing,zhang2018lq,choukroun2019low,dong2019hawq,esser2019learned,nagel2019data,nagel2020up}, there is relatively little work done on quantizing NLP models~\citep{wang2018hitnet,xu2018alternating}, and specifically on transformer models.

Understanding the challenges of transformer quantization and 
designing a robust and easy-to-use quantization pipeline for them
constitute the primary goal of this paper.
The contributions of our work include:
\begin{itemize}
    \item We show that standard 8-bit post-training quantization techniques lead to a significant performance degradation for transformer encoder models.
    \item We conduct a systematic study to identify the underlying reason that precludes efficient transformer quantization.
    We find that the main bottleneck is a considerable mismatch between the different dynamic ranges of activation tensors in the residual connections.
    Further analysis shows that these activation tensors contain structured outliers that facilitate specific attention patterns in deeper encoder layers, such as attending to the special~\texttt{[SEP]} token.
    We highlight that this issue is inherent to many architectures and pre-training objectives.
    \item Based on these findings, we propose a set of solutions with different trade-offs to overcome the dynamic range problem, including techniques based on post-training, mixed precision, and quantization-aware training.
    In particular, we introduce a new~\emph{per-embedding-group} quantization scheme, which solves the activation quantization issue without a significant compute overhead or increase in complexity.
    \item Finally, we show that weights and embeddings in BERT-like models can be quantized to ultra-low (2-4) bits, reducing the memory footprint by more than 8$\times$ with a minimal accuracy loss.
\end{itemize}
We evaluate our proposed solutions on eight different NLP tasks from the well-known GLUE benchmark.
Our techniques set a new state-of-the-art of post-training quantization and per-tensor quantization-aware training for the BERT model.
To the best of our knowledge, this is the first work for the BERT-like transformer quantization with a strong focus on post-training quantization.
The presented method is not exclusive to BERT and is easily applicable to other pre-trained transformer models.


\section{Background and related work}
\label{sec:related_work}

\paragraph{Efficient Transformers}
Making transformer models more efficient in terms of memory and computation time is an active area of research. A good survey paper is~\citet{tay2020efficient}.
%
Most prior work focuses on architectural changes that speed up self-attention, which is the most expensive operation crucial for efficient processing of long sequences of tokens or pixels.
Notable examples include ones that apply fixed~\citep{child2019generating,beltagy2020longformer} or learned~\citep{kitaev2020reformer} sparsity patterns to the otherwise dense attention matrix, while others introduce efficient approximations based on low-rank~\citep{wang2020linformer} or kernel methods~\citep{katharopoulos2020transformers,choromanski2020rethinking}.
%
Some of the complementary efforts in this area are 
compact and fast architectures by design~\citep{sun-etal-2020-mobilebert,iandola-etal-2020-squeezebert},
weight sharing~\citep{dehghani2018universal,lan2019albert},
parameter reuse across multiple downstream tasks~\citep{houlsby2019parameter,stickland2019bert},
knowledge distillation~\citep{sanh2019distilbert,jiao-etal-2020-tinybert},
neural architecture search~\citep{guo2019nat,wang-etal-2020-hat},
pruning~\citep{sanh2020movement,prasanna2020bert}, and
better pre-training~\citep{liu2019roberta,clark2020electra}.

\paragraph{Quantization}
One of the most powerful ways to decrease the computational time and memory consumption of neural networks is quantization, which uses low-bit representations for weight and/or activation tensors.
When moving from 32 to 8 bits, the memory overhead of storing tensors decreases by a factor of 4, while the computational cost for matrix multiplication reduces quadratically by a factor of 16.
Low-bit fixed-point representations, such as INT8, further reduce the energy consumption since the fixed-point operations are more efficient than their floating-point counterparts~\citep{horowitz}.
However, exact latency improvements and energy savings are highly dependent on the target hardware.
Therefore, in this work, we focus on achieving high memory and compute reduction while maintaining acceptable model accuracy and do not measure actual on-device performance gains. We will cover the relevant basics of quantization below. For a more comprehensive overview of neural network quantization, please refer to~\citet{quantization_whitepaper}.

A commonly used scheme for quantization is \emph{uniform affine} or \emph{asymmetric} quantization~\citep{zhou2016dorefa,hubara2017quantized,krishnamoorthi2018quantizing} because it allows for efficient implementation of fixed-point arithmetic.
It is defined by~\emph{bit-width} $b\in\N$, \emph{scale factor} $s\in\R_{+}$, and \emph{zero-point} $z\in\Z$.
We simulate the quantization process in floating-point according to~\citet{jacob2018quantization}.
Quantizing a real-valued tensor $\v{x}$ is performed by first mapping it to an unsigned integer grid:
\begin{equation}
    \v{x}^{\p{\Z}} = \mathrm{clip}\p{\round{\frac{\v{x}}{s}}+z;0,2^b - 1},
    \label{eq:quant_int}
\end{equation}
It is possible to approximately recover the real-valued input $\v{x}$ through an operation that is often referred to as~\emph{de-quantization}:
\begin{equation}
    \wh{\v{x}} := q\p{\v{x};\,s,z,b} = s \p{\v{x}^{\p{\Z}} - z} \approx \v{x}.
    \label{eq:dequant}
\end{equation}
%
In the case of symmetric quantization, we restrict the quantization grid to be symmetric around $z$.

It is common to have a single set of quantization parameters per tensor, known as~\emph{per-tensor quantization}.
One could also increase the ~\emph{quantization granularity} by defining separate quantizers for individual segments of a tensor.
This will improve the accuracy of a network, but at the cost of an additional compute and memory overhead.

An important class of quantization methods is \emph{post-training quantization} (PTQ) algorithms, which take a pre-trained FP32 network and convert it directly into a fixed-point network without the need for the original training pipeline~\citep{krishnamoorthi2018quantizing}.
A vital step in the PTQ process is finding good quantization ranges for each quantizer.
One way of doing this is~\emph{static range estimation}, which determines quantization parameters for the network by passing a few batches of calibration data through the model before inference.
It yields more efficient inference since all the quantization parameters are known in advance and fixed.
Several of the most common range estimators include:
\begin{description}[font=\normalfont\textsl]
    \item[current min-max] or simply \textsl{min-max}, uses the full dynamic range of the tensor~\citep{zhou2016dorefa,wu2018training,zhu2020towards};
    \item[running min-max] uses exponential moving average of the min and max over multiple batches~\citep{krishnamoorthi2018quantizing};
    \item[MSE] finds quantization parameters that minimize mean squared error between quantized and floating-point tensors~\citep{choukroun2019low,banner2018post}.
\end{description}

An alternative to PTQ is to train a neural network with the simulated quantization operations in the network, known as \emph{quantization-aware training} (QAT,~\citealt{jacob2018quantization,gupta2015deep,krishnamoorthi2018quantizing}).
It allows the model to better adapt to the introduced quantization noise compared to PTQ,
at the cost of longer training times, the need for labeled data and doing a hyper-parameter search.
Gradients through the non-differentiable quantization step are usually approximated using the~\emph{straight-through} estimator~\citep{bengio2013estimating}.
Ranges for both weights and activations can be set using PTQ range estimators or learned jointly with the weights during training, as in~\citet{esser2019learned,jain2019trained}.

Finally, it is possible to assign different bit-widths for different layers or parts of the network, a technique known as \emph{mixed precision}~\citep{lin2016fixed,wu2018mixed,zhou2018adaptive,dong2019hawq,wang2019haq,van2020bayesian}.

\paragraph{Transformer quantization}
%
\citet{junczys2018marian} applied knowledge distillation and 8-bit post-training quantization to speed up transformer models for neural machine translation.
\citet{bhandare2019efficient} also applied 8-bit post-training quantization to the transformer model for machine translation and demonstrated how to utilize specialized hardware to accelerate the inference process.

\citet{zafrir2019q8bert} proposed an 8-bit quantization scheme for BERT-like models and achieves compression of up to 25\% of the original model size.
\citet{shen2020q} applies mixed-precision quantization on BERT, where they assign a different precision to different layers according to their sensitivity defined by Hessian information.
\citet{kim2021bert} proposed a fully integer-only arithmetic inference scheme based on second-order polynomial approximations for GELU, Softmax, and LayerNorm non-linearities.
%
Some examples of transformer-based model quantization with alternative quantization schemes include~\citet{fan2020training,NEURIPS2020_747e32ab}.

Note that all the mentioned approaches for BERT-like transformer quantization employ some form of QAT and either do not discuss PTQ alternatives or only use them as weak baselines.


\section{Problem investigation}
\label{sec:problem}

\begin{table*}[htb]
    \centering
    \begin{tabular}{ lccccccccc }
        \toprule
         Configuration & CoLA & SST-2 & MRPC & STS-B & QQP & MNLI & QNLI & RTE & GLUE \\
         \midrule       
         FP32 & 57.27 & 93.12 & 88.36 & 89.09 & 89.72 & 84.91 & 91.58 & 70.40 & \bf{83.06} \\
         \midrule  
         W8A8 & 54.74 & 92.55 & 88.53 & 81.02 & 83.81 & 50.31 & 52.32 & 64.98 & \bf{71.03} \\
         W32A8 & 56.70 & 92.43 & 86.98 & 82.87 & 84.70 & 52.80 & 52.44 & 53.07 & \bf{70.25} \\
         W8A32 & 58.63 & 92.55 & 88.74 & 89.05 & 89.72 & 84.58 & 91.43 & 71.12 & \bf{83.23} \\
         \bottomrule
    \end{tabular}
    \vspace{-.1cm}
    \caption{Post-training quantization results on development sets of the GLUE benchmark (except WNLI). 
The metrics for these tasks can be found in the GLUE paper~\citep{wang-etal-2018-glue}; in all cases, higher is better.
FP32 baseline is trained by the authors from the pre-trained checkpoint, see Appendix~\ref{app:fp32_finetuning} for details.
We report a median over 5 runs with different random seeds.
}
    \label{tbl:03_initial_ptq_results}
\end{table*}


First, we investigate what happens when we apply standard 8-bit post-training quantization to the BERT model and evaluate it on eight downstream tasks from the GLUE benchmark~\citep{wang-etal-2018-glue}.
To quantize fine-tuned models, we use uniform affine quantization with static range estimation, as described in Section~\ref{sec:related_work}.
We quantize all layer's weights and activations (both input and output).
We follow a typical setup with symmetric weight and asymmetric activation quantization~\citep{bhalgat2020lsq+}.
We try several choices for range estimation for both weights and activations and report the best configuration per task, based on its metric (see Appendix~\ref{app:ptq_range_estimators} for details).
In Table~\ref{tbl:03_initial_ptq_results}, we present the results for joint (W8A8), activation-only (W32A8), and weight-only quantization (W8A32).
We note that there is a significant performance degradation for joint 8-bit quantization.
We can also see that weight quantization incurs almost no error on its own and that most degradation is due to activation quantization.
Finally, some tasks seem to be more robust to quantization than others.

To find which part of the network is the most problematic, we perform an ablation study in which we do not quantize specific activations.
The results are summarized in Table~\ref{tbl:03_loo_analysis}.
\begin{table*}[ht]
\begin{floatrow}

    \capbtabbox{%
    \begin{tabular}{ lcccc }
        \toprule
         Quantized activations & STS-B & MNLI & QNLI & RTE \\
         \midrule  
         none (FP32 model) & 89.09 & 84.91 & 91.58 & 70.40 \\
         all & 62.64	&	42.67	&	50.74	&	48.74	\\
         \midrule  
         all, except softmax input & 70.92	&	42.54	&	51.84	&	48.74	\\
         all, except sum of embeddings & 67.57	&	46.82	&	51.22	&	51.26	\\
         all, except self-attention output & 70.47	&	46.57	&	50.98	& 50.90 \\
         all, except softmax output & 72.83	&	50.35	&	50.23	&	49.46	\\
         all, except residual connections after FFN & \bf{81.57}	&	\bf{82.56}	&	\bf{89.73}	&	\bf{67.15}	\\
         same as above, but for layers 10, 11 only & \bf{79.40}	&	\bf{81.24}	&	\bf{88.03}	&	\bf{63.90}	\\
         \bottomrule
    \end{tabular}
    }

    \hspace{-3pt}
    \includegraphics[width=0.19\textwidth]{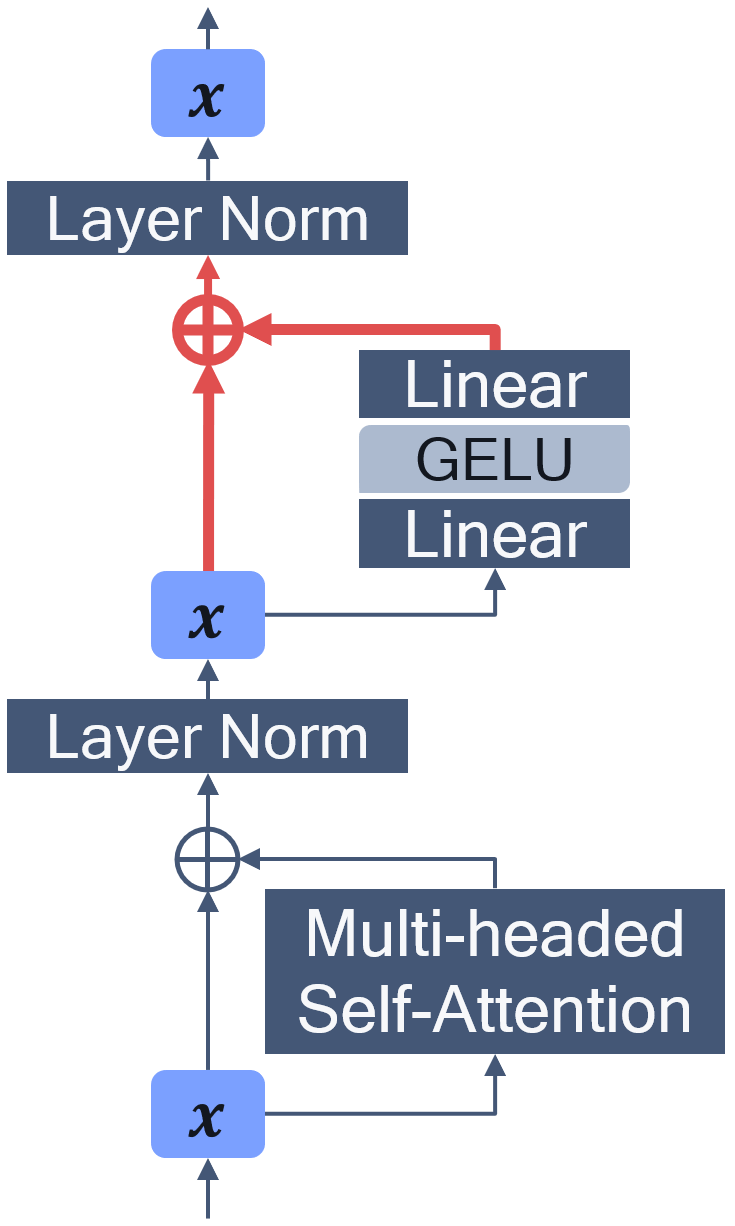}
    \hspace{-6pt}
    
    \captionlistentry[figure]{}
    \label{fig:03_attention_layer}
    \captionsetup{labelformat=andfigure}
    \caption{%
    \emph{Left}: Leave-one-out analysis for activation quantizers on problematic GLUE tasks.
    We set all weights to FP32 and use current min-max (with a batch size of 1) range estimator for activations. 
    We report median score over 5 runs with different random seeds.
    \emph{Right}: A schematic illustration of the attention layer in BERT. 
    Hidden activation tensor is denoted by $\v{x}$. $\oplus$ is an element-wise addition.
    A problematic residual connection sum after feed-forward network is highlighted in red.
    }
    \label{tbl:03_loo_analysis}
\end{floatrow}
\end{table*}

By far, the smallest performance drop is when we do not quantize the residual sum after the feed-forward network (FFN, see Figure~\ref{fig:03_attention_layer}).
Furthermore, the issue seems to be the most pronounced for deeper encoder layers (10 and 11).

To understand why quantizing the residual FFN sum is so detrimental, we look at activation tensors in the problematic 11th layer.
\begin{figure*}[htb]
\subfloat[]{
    \label{fig:03_act_tensors_tokens}
    \includegraphics[width=.34\textwidth]{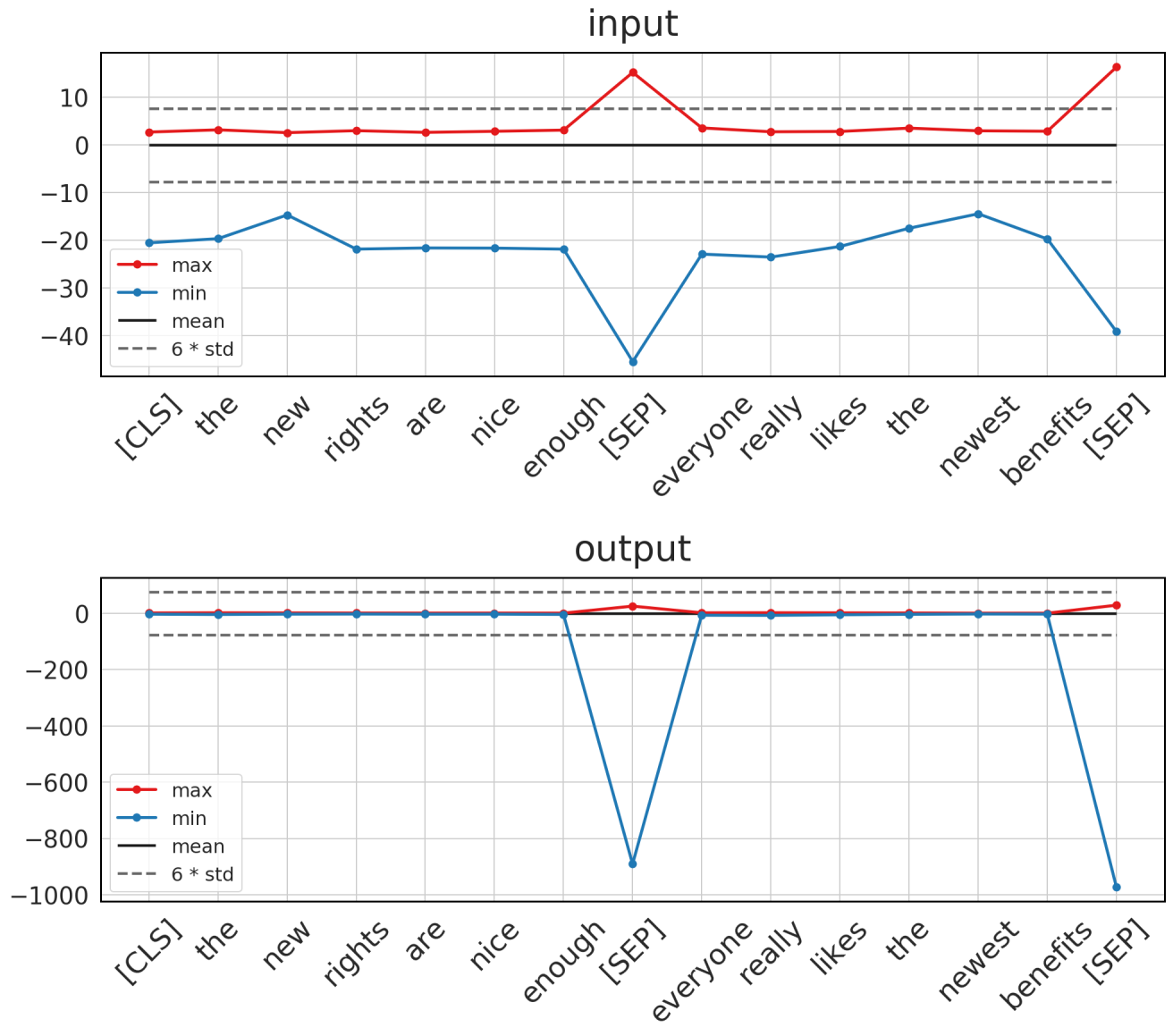}
}
\hspace{13pt}
\subfloat[]{
    \label{fig:03_act_tensors_embd_dims}
    \includegraphics[width=.6\textwidth]{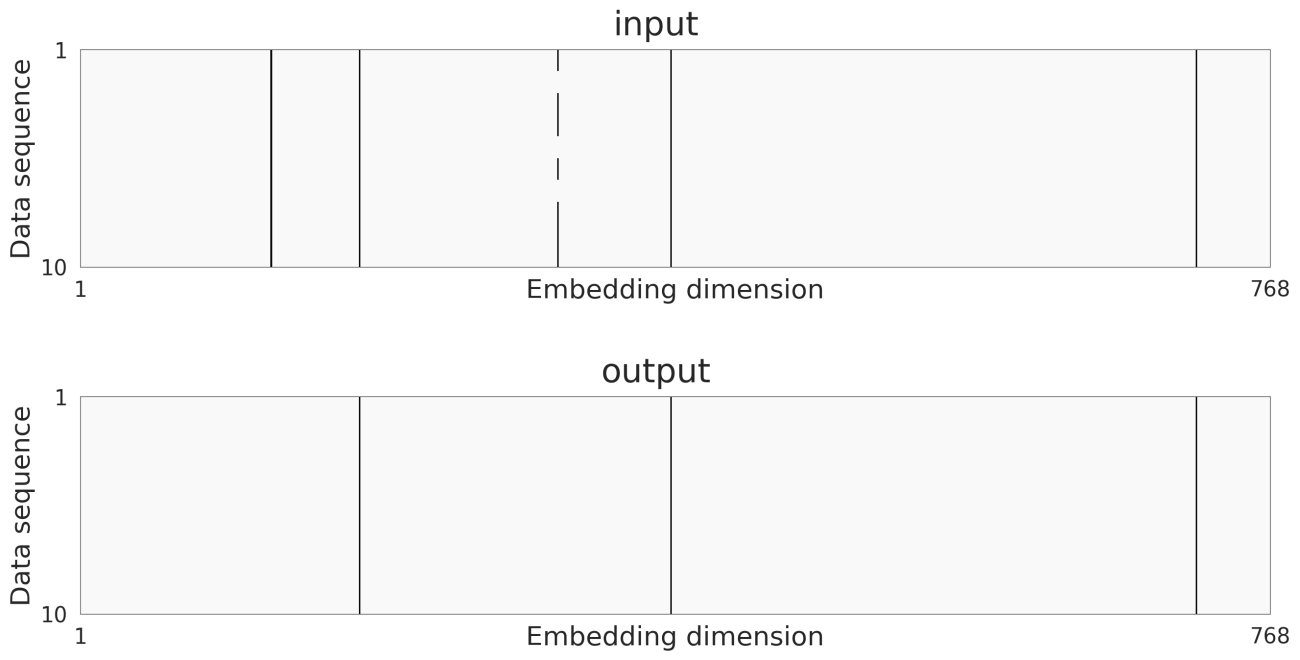}
}
\centering
\vspace{-.1cm}
\caption{%
Full-precision FFN input (top row) and output (bottom row) in 11th layer of BERT.
\protect\subref{fig:03_act_tensors_tokens} Per-token ranges for first data sequence in the MNLI development set.
\protect\subref{fig:03_act_tensors_embd_dims} Visualization of outliers across embedding dimension for the first ten data sequences in the MNLI development set. 
Dark grey color indicates values that exceed six standard deviations from the mean of the activation tensor. 
}
\label{fig:03_act_tensors}
\end{figure*}
%

%
First, from Figure~\ref{fig:03_act_tensors_tokens}, we note that FFN's input and output have radically different dynamic ranges (note the scale for y-axes) due to strong outliers in the output tensor.
Applying per-tensor quantization for the FFN's residual sum is likely to cause a notable error because of the following trade-off between the range and the precision.
On the one hand, using a high dynamic range for small ranged values leads to a loss in representation (high rounding error).
On the other hand, a small dynamic range for large ranged values leads to a very high clipping error.
For such a drastic difference in dynamic ranges, it is very difficult to find the right trade-off between these two kinds of errors.
We also notice a correlation of outliers with special~\texttt{[SEP]} tokens.
In addition to that, from Figure~\ref{fig:03_act_tensors_embd_dims}, we observe that only a few embedding dimensions are consistently responsible for these outliers across many data points.
%

In Appendix~\ref{app:additional_graphs}, we show that this is the case for all layers of BERT-base and all GLUE tasks.
Furthermore, we show that a similar issue is also present in multiple architectures and training objectives, including pre-trained BERT-large, RoBERTa, and DistilRoBERTa~\citep{sanh2019distilbert}, and MobileBERT~\citep{sun-etal-2020-mobilebert}.
%

Further analysis suggests that structured outliers in the FFN's residual connections lead to structured outliers in query-key multiplications in specific attention heads in the next attention layer, causing most of the tokens to attend to the special~\texttt{[SEP]} token.
See Appendix~\ref{app:why_outliers_exist} for more details on this.


\section{Methodology}
\label{sec:method}

In this section, we introduce our proposed techniques for BERT-like model quantization.
Motivated by our findings from Section~\ref{sec:problem}, we consider three ways of efficient BERT quantization -- post-training mixed precision, a new per-embedding-group activation quantization, and quantization-aware training.
Each of these three methods comes with its own set of trade-offs, which is why we present all three. The reader can pick an appropriate solution for their practice. 
As before, we employ uniform affine quantization and static activation ranges, which are either estimated in PTQ or learned during QAT, as described in Section~\ref{sec:related_work}.

\paragraph{Mixed precision PTQ}
As seen in Section~\ref{sec:problem}, not all parts of BERT are equally sensitive to the quantization noise.
Thus, selecting a higher bit-width for sensitive tensors can lead to better accuracy while efficiently keeping all the other tensors in 8-bit or lower.

First, we consider 16-bit activation quantization for problematic activation tensors, such as the residual sum tensor after the feed-forward network.
Higher bit-width will provide a model with sufficient precision to represent both FFN's input and output, as well as their sum.
Additionally, given the observation from Table~\ref{tbl:03_initial_ptq_results}, that the BERT model seems to be quite resilient to 8-bit weight quantization, we also consider the effect of low-bit (2-4) weight and token embedding quantization, which reduces the model size by more than 8$\times$ with a minimal loss in accuracy.

\paragraph{Per-embedding-group PTQ}
As discussed in Section~\ref{sec:related_work}, another way of improving the performance of the quantized model is to increase the quantization granularity.
\begin{figure*}[htb]
\subfloat[Per-tensor]{
\label{fig:04_per_tensor_act_quant}
\includegraphics[width=0.25\textwidth]{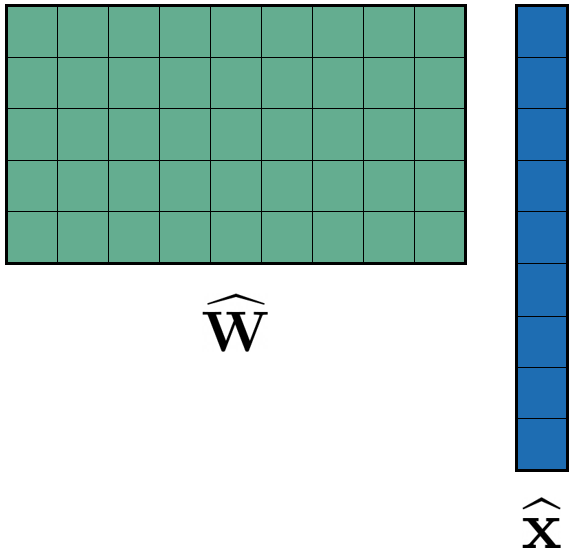}
}
\hfill
\subfloat[Per-embedding]{
\label{fig:04_per_embd_act_quant}
\includegraphics[width=0.25\textwidth]{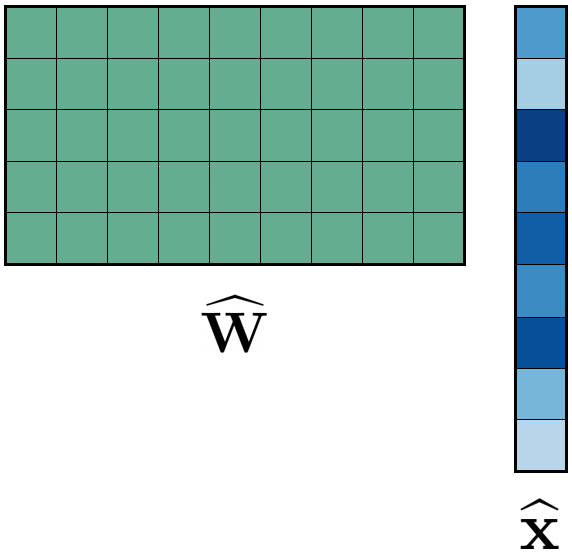}
}
\hfill
\subfloat[Per-embedding-group]{
\label{fig:04_per_embd_group_act_quant}
\includegraphics[width=0.25\textwidth]{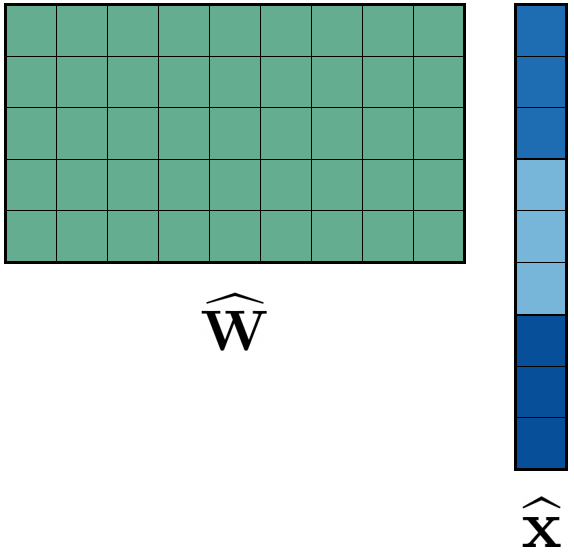}
}
\centering
\caption{An overview for several choices of activation quantization granularity. The color indicates quantization parameter sharing. In all cases we assume per-tensor weight quantization.}
\label{fig:04_act_quant_granularity}
\end{figure*}

Based on our observation from Figure~\ref{fig:03_act_tensors_embd_dims}, that the most problematic outliers in activation tensors are in few designated embedding dimensions, we consider having distinct quantization parameters for individual embedding dimensions or groups of embedding dimensions, as shown in Figure~\ref{fig:04_act_quant_granularity}.

We start by describing \emph{per-embedding} activation quantization.
In BERT-like models, an intermediate hidden activation tensor $\v{x}$ has a shape $(B,T,d)$, where $B$ is the batch size, $T$ is the sequence length, and $d$ is the number of embedding dimensions ($d=768$ for BERT-base, \citealt{devlin-etal-2019-bert}).
Inspired by per-channel weight quantization~\citep{krishnamoorthi2018quantizing}, we can have distinct scaling factors and zero-points per embedding dimension instead of having two scalars for the whole tensor.
In this case, we can collectively denote the quantization parameters by vectors $\v{s},\v{z}\in\R^d$.
The rest of the quantization machinery works as before, including range estimation, with the only difference that equations~\eqref{eq:quant_int} and~\eqref{eq:dequant} are now with broadcasting along the last dimension.
The proposed scheme should alleviate the activation quantization issue since the outlier embedding dimensions will no longer dominate the ranges of other embedding dimensions.

Note, however, that full per-embedding activation quantization will lead to a more expensive computational graph.
To illustrate why, 
consider a matrix-vector multiplication $\m{W}\v{x}$, which in case of per-tensor quantization (and assuming $z=0$) for both weights and activations can be computed as follows:
\begin{flalign}
    \widehat{\m{W}} \widehat{\v{x}} 
    &= \p{s^{\v{w}}\cdot \m{W}^{\p{\Z}}} \p{s^{\v{x}}\cdot \v{x}^{\p{\Z}}} \nonumber \\
    &= s^{\v{w}} s^{\v{x}}\cdot \bigg( \sum_{j=1}^d \m{W}^{\p{\Z}}_{ij} \, \v{x}^{\p{\Z}}_j \bigg)_i.
    \label{eq:per_tensor_matmul}
\end{flalign}
A crucial detail here is that we can factor a common factor $s^{\v{w}} s^{\v{x}}$ out of the summation.
The sum is then efficiently calculated using integer-only arithmetic.
In case of per-embedding activation quantization ($\widehat{\v{x}}=\v{s}^{\v{x}} \odot \v{x}^{\p{\Z}}$), the matrix-vector multiplication becomes instead:
\begin{equation}
    \widehat{\m{W}} \widehat{\v{x}} =
    s^{\v{w}} \cdot  \bigg( \sum_{j=1}^d \v{s}^{\v{x}}_j \cdot \m{W}^{\p{\Z}}_{ij} \, \v{x}^{\p{\Z}}_j \bigg)_i.
    \label{eq:per_embd_act_matmul}
\end{equation}
Here it is no longer possible to take the scaling factor out of the summation and perform a single re-scaling of the result. Instead, one has to perform repeated intermediate re-scalings on the accumulator.

To alleviate the overhead of constant re-scaling, we introduce \emph{per-embedding-group} (PEG) quantization, where we split the activation tensor into $K$ evenly sized groups along the embedding dimension and share quantization parameters among elements in the same group:
\begin{flalign}
    \widehat{\v{x}} = \Big[ \v{s}^{\v{x}}_1 \cdot \b{\v{x}^{\p{\Z}}_1 \ldots \v{x}^{\p{\Z}}_{d/K}}&, \v{s}^{\v{x}}_2 \cdot \b{\v{x}^{\p{\Z}}_{d/K+1}\ldots}, 
    \nonumber \\
    \ldots\,&, \v{s}^{\v{x}}_K \cdot \b{\ldots\v{x}^{\p{\Z}}_{d}}  \Big]
    ,
    \label{eq:per_embd_group_act_quant}
\end{flalign}
where $\b{\cdots}$ denotes concatenation.
Thus the number of required re-scaling operations is significantly reduced from $d$ to $K$.

To ensure all outliers end up in the same group, we employ a deterministic \emph{range-based permutation} of the embedding dimensions.
Similar to range estimation for the activation quantization, we pass some calibration data through the un-quantized network and record the dynamic range $\v{r}_j:=\max(\v{x}_{:,:,j})-\min(\v{x}_{:,:,j})$ for each embedding dimension $j$. 
Next, we define $K$ evenly sized groups based on indices in $\argsort(\v{r})$.
During the range estimation phase, we determine a separate quantization range for each group.
The sorting and grouping need to happen only once before the range estimation phase and deployment to the target.
%

Note that the PEG quantization has a negligible memory overhead, introducing only $d+2\cdot3\cdot K$ extra parameters per attention layer (permutation indices and scale \& zero points per group for FFN's input, output, and sum), which is less than 0.04\% of the total size of BERT-base model.

\paragraph{Simulating PEG quantization using per-tensor operations}
\begin{figure*}[htb]
\includegraphics[width=\textwidth]{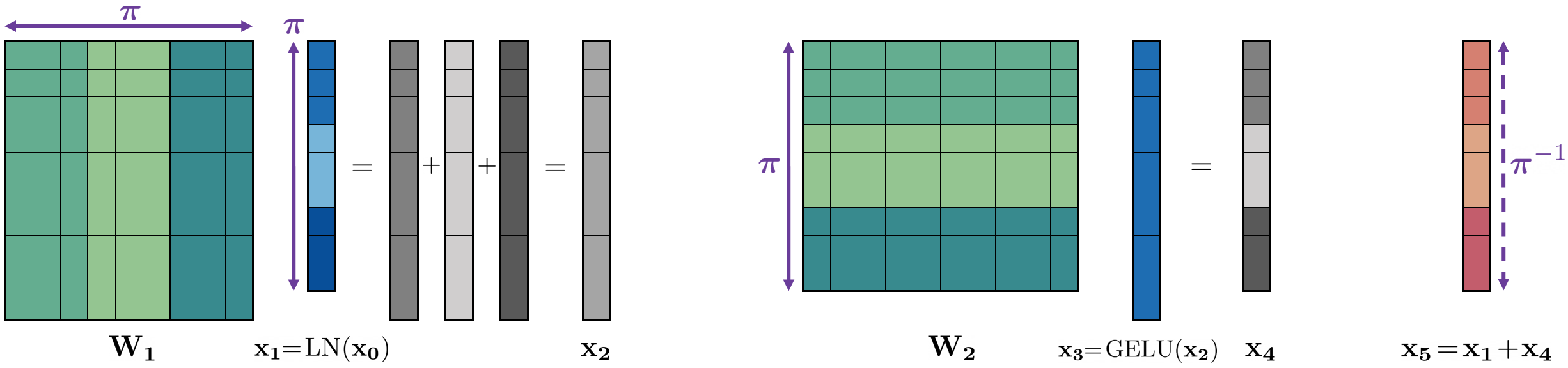}
\centering
\caption{%
A schematic overview of how PEG quantization can be simulated on hardware that only supports per-tensor quantization.
$\v{x_0}$ is the input to the first LayerNorm in the attention layer (see Figure~\ref{fig:03_attention_layer}), $\text{LayerNorm}(\v{x_5})$ is the output of the attention layer, $\m{W_1}$ and $\m{W_2}$ denote weight matrices of linear layers.
The color indicates different embedding groups.
Purple arrows indicate which elements are permuted according to the range-based permutation ($\boldsymbol{\pi}$) or its inverse ($\boldsymbol{\pi}^{-1}$).
}
\label{fig:04_peg_using_per_tensor}
\end{figure*}

%
The proposed scheme might not be natively supported on all target devices, but there is an efficient way to implement it using only per-tensor quantization, illustrated in Figure~\ref{fig:04_peg_using_per_tensor}.
First, consider a case without range-based permutation.
Before quantizing the output of the first LayerNorm (Figure~\ref{fig:03_attention_layer}), we split the output tensor based on embedding groups into $K$ individual tensors.
We also accordingly split columns of the first Linear layer and rows of the second layer and decompose them into $K$ smaller Linear layers each.
The outputs of the first set of layers are elementwise-summed, and the outputs of the second set of layers are concatenated before the residual sum. 
With this functionally equivalent rewriting, all operations can be performed using standard per-tensor quantization.

PEG quantization with permutation can still be simulated on hardware that only supports per-tensor operations.
First, we can share the same permutation for FFN's input, output and sum since we expect the outliers in the output dominate the ones from the input.
Second, we use the permutation-equivariant properties of LayerNorm and linear layers (weights are permuted accordingly before the inference).
We first permute the output of the first LayerNorm, proceed as described above, and then apply inverse permutation before the next LayerNorm.

\paragraph{Quantization-aware training}
Finally, we consider a variant of QAT with learnable ranges for both weights and activations by adapting the procedure from~\citet{esser2019learned,jain2019trained} for BERT-like transformer models.
Simulating the quantization process during fine-tuning allows the model to adapt to quantization noise and often significantly increases performance compared to post-training quantization.

\paragraph{Comparison of methods}
\begin{table}[htb]
    \setlength{\tabcolsep}{5pt}
    \centering
    \begin{tabular}{ lccc }
        \toprule
         Criterion & MP-PTQ & PEG-PTQ & QAT \\
         \midrule       
         Post-training & \checkmark & \checkmark & \ding{53} \\
         Per-tensor & \checkmark & \ding{53} & \checkmark  \\
         Same bit-width & \ding{53} & \checkmark & \checkmark  \\
         \bottomrule
    \end{tabular}
    \caption{Comparison between proposed techniques (MP = mixed precision, PEG = per-embedding-group).}
    \label{tbl:04_method_comparison}
\end{table}

We summarize different trade-offs for the proposed techniques in Table~\ref{tbl:04_method_comparison}.
%
As discussed in Section~\ref{sec:related_work}, usually PTQ methods are preferred over QAT algorithms since they are faster and require either no data or only a small calibration dataset. Additionally, they typically require almost no hyperparameter tuning, enabling easy and computationally efficient quantization.
%
Allocating a higher bit-width to certain parts of the network will reduce the efficiency gain from quantization, because higher bit-width layers are more computationally expensive. It is also not supported by all target hardware.
%
Per-embedding-group quantization has a smaller granularity compared to per-tensor quantization.
It leads to a minor amount of extra compute (and potential latency) due to the additional summation and re-quantization that occurs and might not be supported natively on every fixed-point platform.
Meanwhile, we have shown a way to simulate this scheme on a hardware that only support per-tensor quantization operations.


\section{Experiments}
\label{sec:experiments}

In this section, we evaluate the proposed quantization techniques for the BERT model on GLUE downstream tasks.


\paragraph{Experimental setup}
%
In all experiments, we use uniform affine quantization -- symmetric weights, asymmetric activations -- with the static activation range setting, as discussed in Section~\ref{sec:related_work}.
We quantize all layer's weights and activations.
%
For 8-bit weight quantization, we use the best range settings found in the experiment from Section~\ref{sec:problem}, which can be found in Appendix~\ref{app:ptq_range_estimators}.
However, for low (<8) bit weight and token embedding quantization, we always use the MSE range estimator, as recommended by~\citet{choukroun2019low,banner2018post}.
We set activation ranges based on min and max from a single input sequence.
%
For PTQ experiments, we report the median score over five runs with different random seeds.

For QAT experiments, we initialize all quantization parameters from the PTQ setup described above.
Similarly to full-precision fine-tuning, we use Adam ~\citep{kingma2014adam} and a maximum sequence length of 128, with padding using a special~\texttt{[PAD]} token for shorter sequences.
We use a typical learning rate schedule from the transformer literature~\citep{devlin-etal-2019-bert,liu2019roberta,lan2019albert} -- a linear warmup for the first 10\% of training steps followed by a linear decay to zero.
We perform a hyper-parameter search over the maximum learning rate, batch size, number of epochs, and the self-attention dropout rate for every task and report the best median score over three runs with different random seeds.
For reproducibility, we included more details on the search space and selected hyper-parameters in Appendix~\ref{app:qat_hparams}.

\paragraph{Mixed precision PTQ}
\begin{table}[htb]
    \centering
    \begin{tabular}{ lcccc }
        \toprule
         Method & STS-B & MNLI & QNLI & RTE \\
         \midrule  
         FP32      & 89.09 & 84.91 & 91.58 & 70.40 \\
         W8A8 PTQ  & 79.78 & 45.60 & 51.73 & 64.98 \\
         \midrule  
         MP-PTQ\textsuperscript{*}    & 85.41 & 82.20 & 88.38 & 66.43  \\
         MP-PTQ\textsuperscript{*\dag} & 85.27 & 82.67 & 90.41 & 68.95  \\
         MP-PTQ\textsuperscript{*\dag\ddag} & 88.00 & 82.67 & 90.41 & 68.95 \\
         \bottomrule
    \end{tabular}
    \vspace{-.1cm}
    \caption{Mixed precision post-training quantization results for BERT-base on development sets of the problematic GLUE tasks. 
    \textsuperscript{*}Uses 16-bit residual FFN sum.
    \textsuperscript{\dag}Uses 16-bit FFN input and output.
    \textsuperscript{\ddag}Uses 16-bit final output (using MSE range estimator).
    }
    \label{tbl:05_mp_ptq_results}
\end{table}

First, we present the results for mixed precision post-training quantization (MP-PTQ) in Table~\ref{tbl:05_mp_ptq_results}, where we start from 8-bit activations and progressively keep more and more operations in 16-bit precision.
We see that for classification tasks (MNLI, QNLI, RTE), it is sufficient to keep a few of the most problematic parts in 16-bit to get good performance.
For the STS-B regression task, it is also necessary to keep the output in higher precision to close the gap with FP32 model performace.

In conclusion, by only keeping 22\% of the activations in 16-bit\footnote{36 out of 161 activation quantizers for BERT-base}, we can achieve performance close to FP32,
while all other activations and all weights are in 8-bit for efficient inference.

\paragraph{Per-embedding-group PTQ}
Next, we investigate the effectiveness of the proposed per-embedding-group post-training activation quantization, depending on the number of groups $K$.
\begin{table}[t]
    \setlength{\tabcolsep}{4pt}
    \centering
    \begin{tabular}{ lcccc }
      \toprule
       \#groups, $K$ & STS-B & MNLI & QNLI & RTE \\
       \midrule  
       FP32 & 89.09 & 84.91 & 91.58 & 70.40 \\
       1 \footnotesize{(=\,per-tensor)} & 79.78 & 45.60 & 51.73 & 64.98 \\
       \midrule  
       768 \footnotesize{(=\,per-embd.)}  & 87.87 & 80.97 & 90.66 & 69.31 \\
       768 \footnotesize{(only FFN)}\textsuperscript{*} & 87.92 & 81.00 & 90.68 & 68.59 \\
       6 \footnotesize{(only FFN)}   & 87.26 & 80.51 & 89.82 & 68.59 \\
       3 \footnotesize{(only FFN)}   & 85.96 & 76.43 & 80.74 & 66.06 \\
       \midrule  
       3\,+\,P \footnotesize{(only FFN)} & \u{87.92} & 80.64 & \u{91.07} & \u{69.31} \\
       6\,+\,P \footnotesize{(only FFN)} & \u{87.92} & \u{81.25} & \u{91.07} & \u{69.31} \\
       \bottomrule
    \end{tabular}
    \vspace{-.1cm}
    \caption{Per-embedding-group activation quantization PTQ results for BERT-base on development sets of the problematic GLUE tasks.
    \textsuperscript{*}Per-embedding-group quantization is applied only to FFN's input, output, and residual sum (all the rest -- per-tensor).
    ``\,+\,P'' -- Uses range-based permutation.
    }
    \label{tbl:05_per_embd_ptq_results}
\end{table}

The results are summarized in Table~\ref{tbl:05_per_embd_ptq_results}.
Per-embedding activation quantization significantly improves performance, even when only applied to problematic parts of the network.
Surprisingly, we can also recover most of the performance degradation with only $K=3$ groups (size 256 each), especially if we apply range-based permutation to ensure all the outliers end up in the same group.
A small number of groups is essential since it limits the number of re-scalings required, enabling efficient execution on resource constraint devices.

\paragraph{Comparison of proposed methods}
\begin{table*}[htb]
    \setlength{\tabcolsep}{4.5pt}
    \centering
    \begin{tabular}{ lccccccccc }
        \toprule
         Method & CoLA & SST-2 & MRPC & STS-B & QQP & MNLI & QNLI & RTE & GLUE \\
         \midrule       
         FP32 baseline & 57.27 & 93.12 & 88.36 & 89.09 & 89.72 & 84.91 & 91.58 & 70.40 & \bf{83.06} \\
         \midrule  
         Our W8A8 PTQ & 54.74 & 92.55 & 88.53 & 81.02 & 83.81 & 50.31 & 52.32 & 64.98 & \bf{71.03} \\
         Our W8A\{8,16\} MP-PTQ & 58.63 & 92.66 & 88.74 & 88.00 & 89.40 & 82.67 & 90.41 & 68.95 & \bf{82.43} \\
         Our W8A8 PEG-PTQ & 59.43 & 92.66 & 88.53 & 87.92 & 89.42 & 81.25 & 91.07 & 69.31 & \bf{82.45} \\
         Our W8A8 QAT & 61.27 & 93.00 & 88.80 & 88.95 & 89.44 & 83.74 & 90.48 & 70.40 & \bf{83.26} \\
         \midrule  
         Q8BERT W8A8 PTQ\textsuperscript{*\dag} & 56.74 & 91.04 & 87.88\textsuperscript{\ddag} & 87.66\textsuperscript{\ddag} & 84.98\textsuperscript{\ddag} & -- & 89.34 & 63.32 & \bf{80.13}\textsuperscript{\S} \\
         Q8BERT W8A8 QAT\textsuperscript{*} & 58.48 & 92.24 & 89.56\textsuperscript{\ddag} & 89.04\textsuperscript{\ddag} & 87.96\textsuperscript{\ddag} & -- & 90.62 & 68.78 & \bf{82.38}\textsuperscript{\S} \\
         Q-BERT W8A8 QAT\textsuperscript{$\psi$} & -- & 92.88 & -- & -- & -- & 83.87 & -- & -- & \bf{--} \\
         \bottomrule
    \end{tabular}
    \vspace{-.1cm}
    \caption{8-bit quantization results for BERT-base on development sets of the GLUE benchmark (except WNLI).
    The metrics for these tasks can be found in the GLUE paper~\citep{wang-etal-2018-glue}; in all cases, higher is better.
    %
    We compare against Q8BERT~\citep{zafrir2019q8bert} and Q-BERT~\citep{shen2020q}.
    Note that these papers start from FP32 baselines with slightly different scores.
    \textsuperscript{*}Uses FP32 Softmax, GELU and LayerNorm.
    \textsuperscript{\dag}Uses dynamic activation quantization.
    \textsuperscript{\ddag}Reports F1 score for MRPC, QQP and Pearson Correlation for STS-B, instead of the combined metrics.
    \textsuperscript{\S}A macro-average without a score for the MNLI task.
    \textsuperscript{$\psi$}Uses group-wise per-channel weight quantization with 128 groups and keeps the last fully-connected layer in FP32.
    }
    \label{tbl:05_main_results}
\end{table*}

We summarize the results for all of our proposed techniques and compare them to several related methods from the literature in Table~\ref{tbl:05_main_results}.
%
We use the same setup as described above.
Unless otherwise stated, all results use 8-bit per-tensor quantization for both weights and activations.
For mixed precision (MP-PTQ), we use the best setup from the ablation study before.
For per-embedding-group quantization (PEG-PTQ), we use $K=6$ groups with range-based permutation for all tasks and only apply it to FFN's input, output, and the sum.

To summarize, all the proposed techniques solved the dynamic range problem, enabling efficient transformer quantization with minimum accuracy loss.
Our PTQ results strongly outperform results from the literature, while our assumptions in mixed precision are milder than ones of Q8BERT, which keeps all non-linearities in FP32.
Our per-tensor QAT results are also on par or outperform results from the literature, which uses finer quantization granularity and keeps certain parts of the network in FP32.

%
\paragraph{Low-bit weight and token embeddings}
\begin{table}[t]
    \centering
    \setlength{\tabcolsep}{4.5pt}
    \begin{tabular}{ lcc }
        \toprule
         \multicolumn{1}{p{4.25cm}}{\centering Method} & \multicolumn{1}{p{1.5cm}}{\centering Memory \\ reduction} & GLUE \\
         \midrule       
         FP32 baseline & $\times$1.00 & 83.06 \\
         \midrule  
         W6A32 PTQ & $\times$5.33 & 81.41 \\
         W4A32 PTQ & $\times$8.00 & 72.31 \\
         W4A32 AdaRound (PTQ) & $\times$8.00 & 81.46 \\
         \midrule
         W4A32 QAT & $\times$8.00 & 82.95 \\
         W4A8 QAT & $\times$8.00 & 82.64 \\
         W4A8, 2-bit embd. QAT & $\times$8.85 & 82.29 \\
         \bottomrule
    \end{tabular}
    \vspace{-.1cm}
    \caption{Low-bit weight \& token embedding quantization results for for BERT-base on development sets of the GLUE benchmark.
    For AdaRound optimization (\citealt{nagel2020up}, our impl.), we used 1024 random data sequences and $10^4$ iterations with default hyper-parameters from the paper.
    }
    \vspace{-.1cm}
    \label{tbl:05_lowbit_weights_embd}
\end{table}

Given the robustness of the BERT model to 8-bit weight quantization,
we investigate the effect of low-bit weight and token embedding quantization and summarize the results in Table~\ref{tbl:05_lowbit_weights_embd}.

We see that even in the post-training regime, it is possible to achieve low-bit weight quantization with acceptable performance degradation, especially when combined with AdaRound~\citep{nagel2020up}, a technique for learning optimal rounding.
%
QAT recovers most of the performance, even with quantized activations. Furthermore, we can push token embeddings to 2-bits with less than a 0.8\% drop in terms of the GLUE score.
It reduces the model size by \textbf{8.85$\times$} compared to the original FP32 checkpoint and can significantly increase inference speed and reduce the energy consumption on resource constraint devices.
%
More detailed results, including per-task scores and comparison to results from the literature, can be found in Appendix~\ref{app:lowbit_weights_embd_full}.


\section{Conclusions}
\label{sec:conclusions}


In this paper, we explored quantization for BERT-like transformers.
%
We showed that these models have unique quantization challenges -- namely, high dynamic activations ranges that are difficult to represent with a low bit fixed-point format.
These activations contain structured outliers in the residual connections that encourage specific model behavior, such as attending to the special~\texttt{[SEP]} token.
%
Motivated by our findings, we proposed three solutions, one based on mixed precision quantization, a novel per-embedding-group quantization, and quantization-aware training.
Each of these methods has its own set of trade-offs in terms of accuracy, ease of use, and model size.
%
Our techniques overcome the dynamic range issues and set a new state-of-the-art for PTQ and per-tensor QAT on GLUE downstream tasks.
Finally, we achieved 4-bit weight and 2-bit token embedding quantization with less than 0.8\% drop in terms of GLUE score, leading to significant memory and compute savings.
%

\bibliography{anthology,references}
\bibliographystyle{acl_natbib}


\clearpage
\newpage

\appendix

\begin{strip}
{\Large\textbf{Supplementary materials}}
\end{strip}
\section{Why do these outliers exist?}
\label{app:why_outliers_exist}
%
\begin{figure*}[htb]
\subfloat[]{
    \includegraphics[width=.175\textwidth]{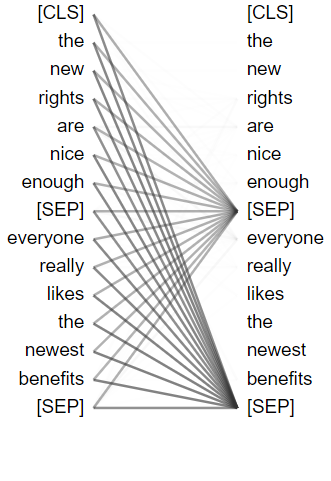}
    \label{fig:A_head}
}
\hfill
\subfloat[]{
    \includegraphics[width=.79\textwidth]{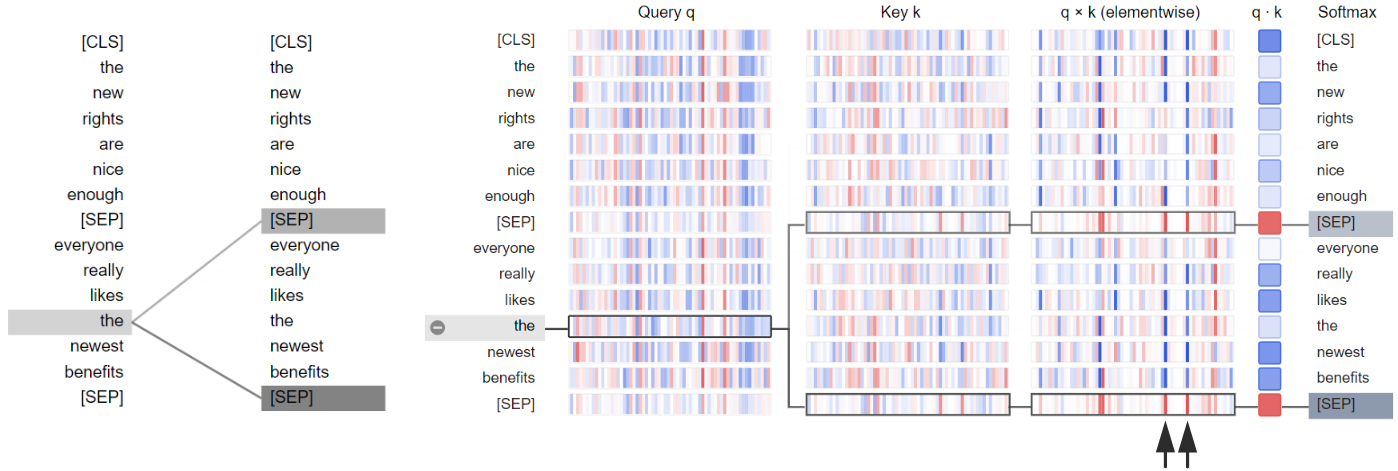}
    \label{fig:A_qkmul}
}
\caption{%
Visualization of the attention pattern in 6th attention head of 11th layer in BERT-base, computed on first data sequence from the MNLI development set.
\protect\subref{fig:A_head} visualization of the attention weights (attention probabilities) between the tokens. Thickness of line between the query vector $\v{q}_i$ of $i$-th token on the left and key vector $\v{k}_j$ of $j$-th token on the right is proportional to $\exp\p{\frac{\v{q}_i \cdot \v{k}_j}{\sqrt{d}}}$. To generate this figure, we used ``head view'' from the BertViz library~\citep{vig-2019-multiscale}.
\protect\subref{fig:A_qkmul} a ``deconstruction'' of the behavior on the left in terms of elementwise query-key vector multiplications. Values in red indicate high positive values, while values in blue indicate negative values. We used ``neuron view'' from the BertViz library to generate this figure.
}
\label{fig:A}
\end{figure*}

%
To better understand why transformer models learn these peculiar outliers, 
we look at what happens with those outliers when they proceed to the next attention layer.
We visualize the attention mechanism for one of the attention heads in the problematic 11th layer of BERT-base in Figure~\ref{fig:A}.
We can see that most of the tokens in this attention head attend to special~\texttt{[SEP]} tokens.
Furthermore, from Figure~\ref{fig:A_qkmul} we see a similar consistent vertical pattern (indicated by black arrows) as we saw from the per-embedding graphs for FFN's input and output activation tensors (see Figure~\ref{fig:03_act_tensors_embd_dims} in paper).
It means the attention mechanism generates such queries and key vectors that the decision of attending to special separator tokens is determined by only a few designated neurons.
It suggests that structured outliers in residual connections lead to structured outliers in query-key multiplications, causing most tokens to attend to the separator token.

\citet{clark-etal-2019-bert} has shown that in BERT-like transformer models, attending to the special~\texttt{[SEP]} token is essentially a ``no-op'' for attention heads that cannot extract patterns they were trained to look for from the specific passage of text. 
\citet{clark-etal-2019-bert} also showed that such behavior is quite common: often, more than a half of the head's attention is on special tokens, specifically in deeper layers.

We hypothesize that such an attention pattern seems to be a useful one to obtain a good predictive performance, while the structured outliers merely help to facilitate this behavior.
These outliers causing such a high dynamic range for activations likely emerged as a result of specific architectural choices (e.g., large fully-connected layers) and long pre-training with no explicit activation regularization applied.


\section{Experimental details}
%
\subsection{FP32 fine-tuning details}
\label{app:fp32_finetuning}
We use pre-trained checkpoint for BERT-base (uncased, 109M parameters) from HuggingFace repository~\citep{wolf-etal-2020-transformers}.
We follow a standard fine-tuning practices from~\citep{devlin-etal-2019-bert} and~\url{https://github.com/huggingface/transformers}.
Each data point is tokenized and truncated to the maximum sequence length of 128. Shorter sequences are padded to the same length of 128 using a special~\texttt{[PAD]} token.
We fine-tune for 3 epochs using Adam for all tasks.
Learning rate is initially set to its maximum value and is linearly decayed to zero by the end of fine-tuning.
We tune the batch size and maximum value of learning rate individually per task from the following search space:
\begin{itemize}
  \item batch size: \{32, 64\} for bigger tasks (QQP, MNLI, QNLI) and \{8, 16, 32, 64\} for the rest,
  \item learning rate: \{2,3,4,5\}e-5.
\end{itemize}
We repeat every experiment 5 times with different random seeds and select the configuration with the best median score on the development set for the respective task.
\begin{table}[htb]
    \centering
    \begin{tabular}{ lcc }
        \toprule
         Task & Learning rate & Batch size \\
         \midrule       
         CoLA  & 2e-05 & 32 \\
         SST-2 & 2e-05 & 16 \\
         MRPC  & 5e-05 &  8 \\
         STS-B & 4e-05 & 32 \\
         QQP   & 4e-05 & 64 \\
         MNLI  & 2e-05 & 16 \\
         QNLI  & 2e-05 & 32 \\
         RTE   & 3e-05 &  8 \\
         \bottomrule
    \end{tabular}
    \caption{Hyper-parameters for FP32 BERT-base fine-tuning on GLUE downstream tasks.}
    \label{tbl:A_hparams_fp32_finetuning}
\end{table}

These configurations are shown in Table~\ref{tbl:A_hparams_fp32_finetuning}.
Quantization is always applied to the median checkpoint for the respective task.

We exclude the problematic WNLI task~\citep{levesque2012winograd}, as it has relatively small dataset and shows an unstable behaviour~\citep{dodge2020fine}, in particular due to several issues with the way the dataset was constructed\footnote{See~\url{https://gluebenchmark.com/faq} for details.}.


\subsection{Range setting for 8-bit post-training quantization}
\label{app:ptq_range_estimators}
We select the best range estimators from the following search space:
\begin{itemize}
  \item weights: \{min-max, MSE\};
  \item activations: \{current min-max, running min-max, MSE\}.
\end{itemize}
For activations, we also select the best batch size and number of batches from \{1,4,16\} (except current min-max, for which only a single batch is used).
For running min-max, we use the momentum coefficient of 0.9.
We repeat every experiment 5 times with different random seeds and select the configuration with the best median score on the development set for the respective task.
\begin{table}[htb]
    \centering
    \begin{tabular}{ lll }
        \toprule
         Task & Weights & Activations (bs, nb) \\
         \midrule       
         CoLA  & min-max & running min-max (1, 4) \\
         SST-2 & MSE     & running min-max (4, 16) \\
         MRPC  & MSE     & running min-max (16, 16) \\
         STS-B & min-max & running min-max (1, 16) \\
         QQP   & min-max & running min-max (16, 16) \\
         MNLI  & min-max & running min-max (1, 16) \\
         QNLI  & min-max & running min-max (1, 16) \\
         RTE   & MSE     & current min-max (1) \\
         \bottomrule
    \end{tabular}
    \caption{Best range estimators for post-training quantization of BERT-base on GLUE tasks (bs = batch size, nb = number of batches).}
    \label{tbl:A_ptq_range_estimators}
\end{table}

Best configurations for joint weight and activation 8-bit post-training quantization are listed in Table~\ref{tbl:A_ptq_range_estimators}.


\subsection{W8A8 QAT hyper-parameters}
\label{app:qat_hparams}
\begin{table}[htb]
    \setlength{\tabcolsep}{4pt}
    \centering
    \begin{tabular}{ lcccc }
        \toprule
         Task  & LR & BS & E & D \\
         \midrule       
         CoLA  & \{\u{2}, 3, 4\}e-5 & \{\u{16}, 32\} & \{\u{3}, 6\} & \{\u{0}, 0.1\}  \\
         SST-2 & \{1, \u{2}, 3\}e-5 & \{16, \u{32}\} & \{\u{3}, 6\} & \{0, \u{0.1}\}  \\
         MRPC  & \{\u{1}, 2, 3\}e-5 &  \{\u{8}\}     & \{\u{3}, 6\} & \{0, \u{0.1}\}  \\
         STS-B & \{2, \u{4}, 8\}e-5 & \{\u{16}, 32\} & \{3, \u{6}\} & \{0, \u{0.1}\}  \\
         QQP   & \{\u{4}, 5, 6\}e-5 & \{\u{32}\}     & \{\u{3}\}    & \{0, \u{0.1}\}  \\
         MNLI  & \{\u{2}, 3, 4\}e-5 & \{\u{16}, 32\} & \{\u{3}\}    & \{0, \u{0.1}\}  \\
         QNLI  & \{2, \u{3}, 4\}e-5 & \{16, \u{32}\} & \{\u{3}\}    & \{\u{0}, 0.1\}  \\
         RTE   & \{1, \u{3}, 5\}e-5 & \{\u{8}\}      & \{3, \u{6}\}  & \{0, \u{0.1}\} \\
         \bottomrule
    \end{tabular}
    \caption{Hyper-parameters for 8-bit QAT for BERT-base on development sets of the GLUE benchmark (LR = maximum learning rate, BS = batch size, E = number of epochs, D = self-attention dropout rate).
    Values in bold indicate the best configuration.
    }
    \label{tbl:A_qat_hparams}
\end{table}

Hyper-parameters for W8A8 quantization-aware training are listed in Table~\ref{tbl:A_qat_hparams}.


\subsection{W4A8 QAT hyper-parameters}
\label{app:w4a8_qat_hparams}
\begin{table}[htb]
    \centering
    \begin{tabular}{ lcccc }
        \toprule
         Task  & LR & BS & E  \\
         \midrule       
         CoLA  & \{\u{2}, 3, 4\}e-5 & \{\u{16}, 32\} & \{\u{3}, 6\}  \\
         SST-2 & \{2, \u{3}, 4\}e-5 & \{16, \u{32}\} & \{\u{3}, 6\}  \\
         MRPC  & \{\u{2}, 3, 4\}e-5 &  \{\u{8}\}     & \{\u{3}, 6\}  \\
         STS-B & \{4, \u{6}, 8\}e-5 & \{16, \u{32}\} & \{3, \u{6}\}  \\
         QQP   & \{4, \u{5.5}, 7\}e-5 & \{\u{32}\}     & \{\u{3}, 6\}     \\
         MNLI  & \{3, \u{4}, 5\}e-5 & \{\u{16}, 32\} & \{\u{3}, 6\}     \\
         QNLI  & \{\u{2}, 3, 4\}e-5 & \{\u{16}, 32\} & \{\u{3}, 6\}     \\
         RTE   & \{3, 4, \u{5}\}e-5 & \{\u{8}\}      & \{3, \u{6}\}  \\
         \bottomrule
    \end{tabular}
    \caption{Hyper-parameters for W4A8 QAT for BERT-base on development sets of the GLUE benchmark (LR = maximum learning rate, BS = batch size, E = number of epochs).
    Values in bold indicate the best configuration.
    }
    \label{tbl:A_qat_lowbit_weights_hparams}
\end{table}

Hyper-parameters for W4A8 quantization-aware training are listed in Table~\ref{tbl:A_qat_lowbit_weights_hparams}.


\section{Detailed results for low-bit weight and embedding quantization}
\label{app:lowbit_weights_embd_full}
\begin{table*}[htb]
    \setlength{\tabcolsep}{3.5pt}
    \centering
    \begin{tabular}{ lccccccccc }
        \toprule
         Method & CoLA & SST-2 & MRPC & STS-B & QQP & MNLI & QNLI & RTE & GLUE \\
         \midrule       
         FP32 baseline & 57.27 & 93.12 & 88.36 & 89.09 & 89.72 & 84.91 & 91.58 & 70.40 & \bf{83.06} \\
         \midrule
         Our W8A32, 6-bit embd. PTQ & 58.65 & 92.32 & 88.36 & 89.06 & 89.74 & 84.62 & 91.32 & 68.95 & \bf{82.88} \\
         Our W8A32, 4-bit embd. PTQ & 57.43 & 92.32 & 88.79 & 89.02 & 89.70 & 84.58 & 91.43 & 68.59 & \bf{82.73} \\
         Our W8A32, 2-bit embd. PTQ & 57.22 & 92.32 & 86.99 & 88.87 & 89.63 & 84.45 & 91.47 & 68.23 & \bf{82.40} \\
         \midrule  
         Our W6A32 PTQ & 56.23 & 91.86 & 86.70 & 87.76 & 88.94 & 82.39 & 88.83 & 68.23 & \bf{81.41} \\
         Our W4A32 PTQ & 43.06 & 90.83 & 84.90 & 83.07 & 79.37 & 68.16 & 79.68 & 50.18 & \bf{72.31} \\
         Our W4A32 AdaRound (PTQ) & 54.56 & 92.32 & 87.53 & 87.91 & 88.30 & 81.61 & 90.17 & 69.31 & \bf{81.46} \\
         \midrule
         Our W4A32 QAT & 58.31 & 92.49 & 87.58 & 89.02 & 89.78 & 84.29 & 91.40 & 70.76 & \bf{82.95} \\
         Our W4A8 QAT & 57.22 & 92.32 & 87.77 & 89.13 & 89.64 & 83.69 & 91.29 & 70.04 & \bf{82.64} \\
         Our W4A8, 2-bit embd. QAT & 56.08 & 91.74 & 87.59 & 89.19 & 89.56 & 83.68 & 90.79 & 69.67 & \bf{82.29} \\
         \midrule
         Q-BERT W4A8 QAT\textsuperscript{*} & -- & 85.67 & -- & -- & -- & 76.85 & -- & -- & \bf{--} \\
         Q-BERT W4A8 QAT\textsuperscript{*\dag} & -- & 92.66 & -- & -- & -- & 84.03 & -- & -- & \bf{--} \\
         \bottomrule
    \end{tabular}
    \caption{%
    Low-bit weight and token embedding quantization results for BERT-base on development sets of the GLUE benchmark.
    We compare against Q-BERT~\citep{shen2020q}.
    Note that this work starts from FP32 baselines with slightly different scores.
    \textsuperscript{*}Keeps the last fully-connected layer in full precision.
    \textsuperscript{\dag}Uses group-wise per-channel weight quantization with 128 groups (of size 6 each).
    }
    \label{tbl:B_lowbit_weights_embd_full}
\end{table*}
         

%
Detailed results for low-bit weight and token embedding quantization for BERT-base on development sets of the GLUE benchmark (including per-task scores) are summarized in Table~\ref{tbl:B_lowbit_weights_embd_full}.
%

\section{Additional graphs from problem investigation}
\label{app:additional_graphs}
%
%
\subsection{Per-embedding outliers for all layers in BERT-base}
\begin{figure*}[htb]
    \includegraphics[width=\textwidth]{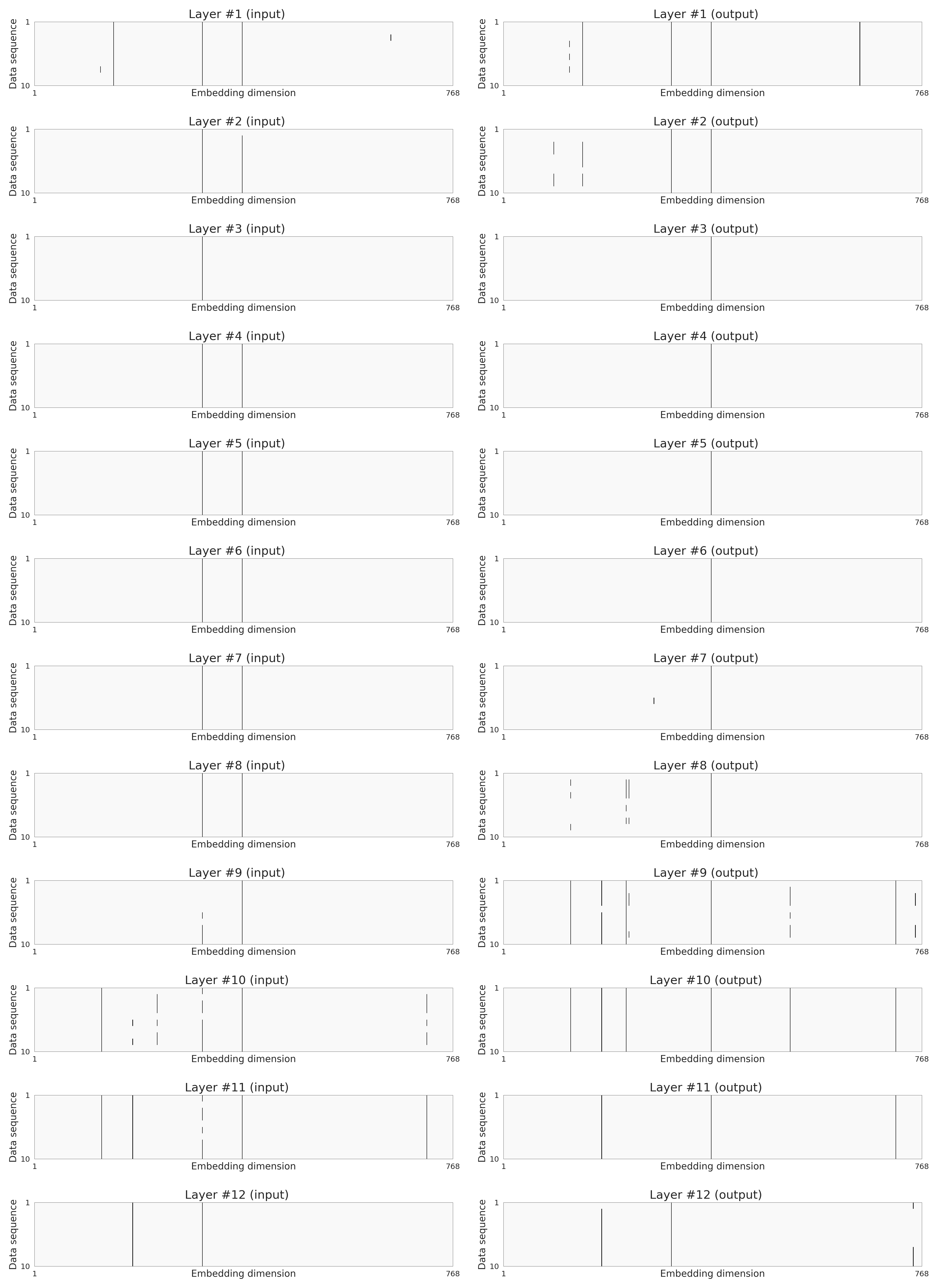}
    \caption{Visualization of activation tensor outliers in BERT-base FFN's input and output across embedding dimension for the first ten data sequences in the MNLI development set.
Dark grey color indicates values that exceed six standard deviations from the mean of the activation tensor.}
    \label{fig:D_mnli}
\end{figure*}

\begin{figure*}[htb]
    \includegraphics[width=\textwidth]{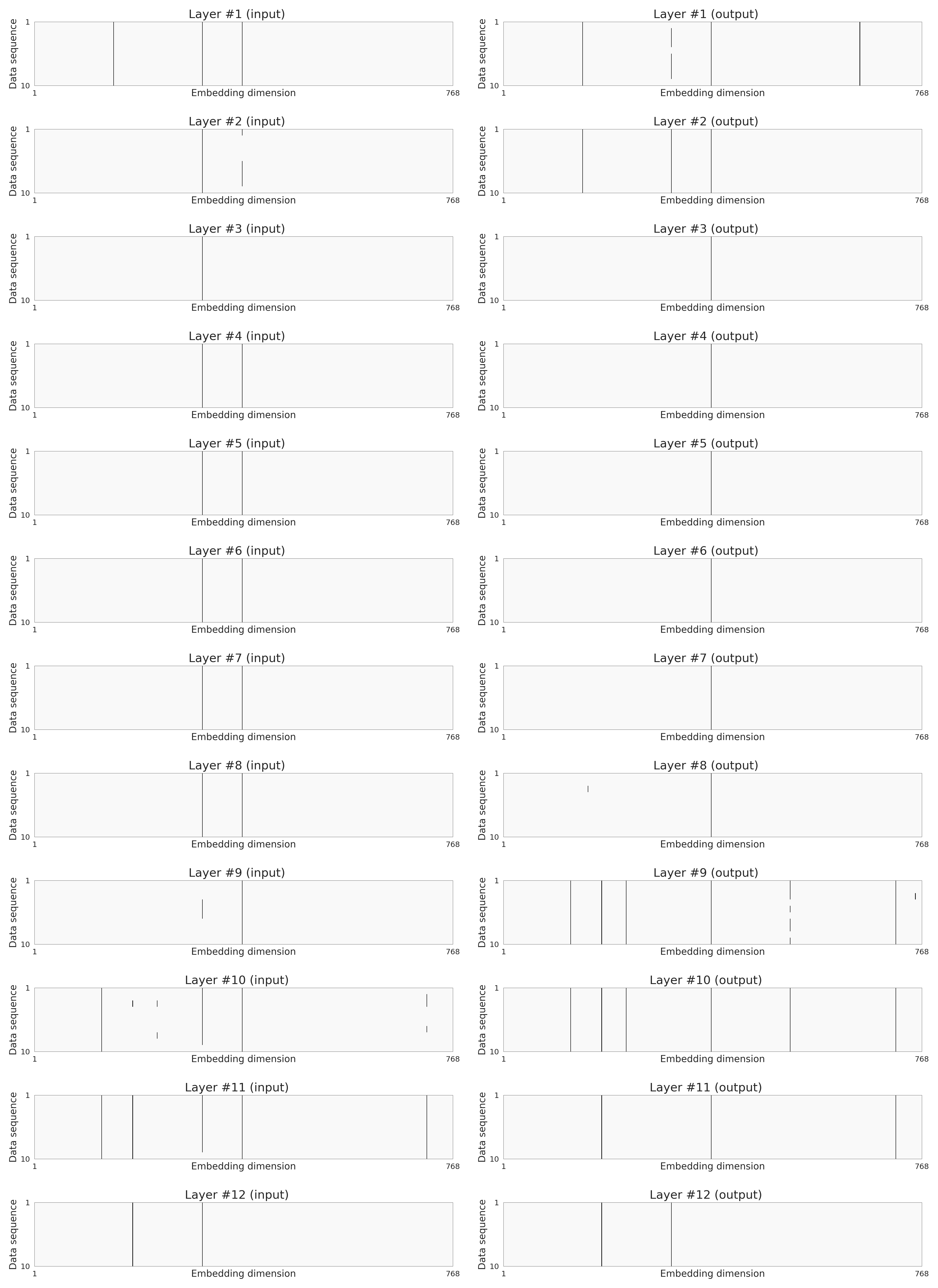}
    \caption{Visualization of activation tensor outliers in BERT-base FFN's input and output across embedding dimension for the first ten data sequences in the STS-B development set.
Dark grey color indicates values that exceed six standard deviations from the mean of the activation tensor.}
    \label{fig:D_stsb}
\end{figure*}

\begin{figure*}[htb]
    \includegraphics[width=\textwidth]{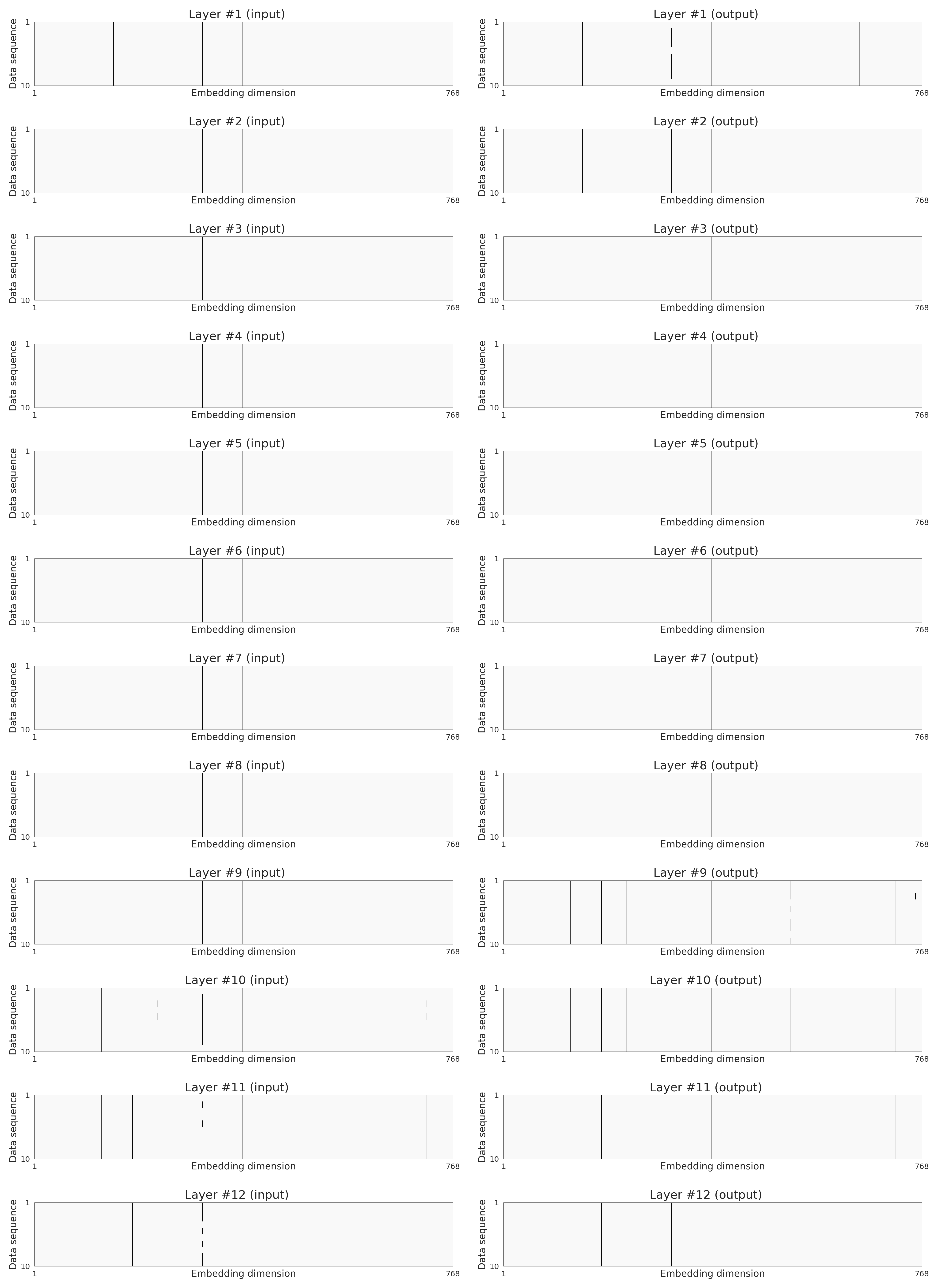}
    \caption{Visualization of activation tensor outliers in BERT-base FFN's input and output across embedding dimension for the first ten data sequences in the MRPC development set.
Dark grey color indicates values that exceed six standard deviations from the mean of the activation tensor.}
    \label{fig:D_mrpc}
\end{figure*}

We visualize per-embedding outliers in FFN's input and output for all layers in BERT-base computed on first ten data sequences from the development set of MNLI (Figure~\ref{fig:D_mnli}), STS-B (Figure~\ref{fig:D_stsb}) and MRPC (Figure~\ref{fig:D_mrpc}).
We see that only a few designated embedding dimensions generate outliers across many data points.
It suggests that such behavior is already pre-determined by the weights and embeddings of the pre-trained BERT model.
%
%
%
%
\subsection{Activation tensors for different architectures}
We shot that dynamic range issue is present in multiple architectures and training objectives:
%
\begin{figure*}[htb]
\subfloat[input]{
    \includegraphics[width=.48\textwidth]{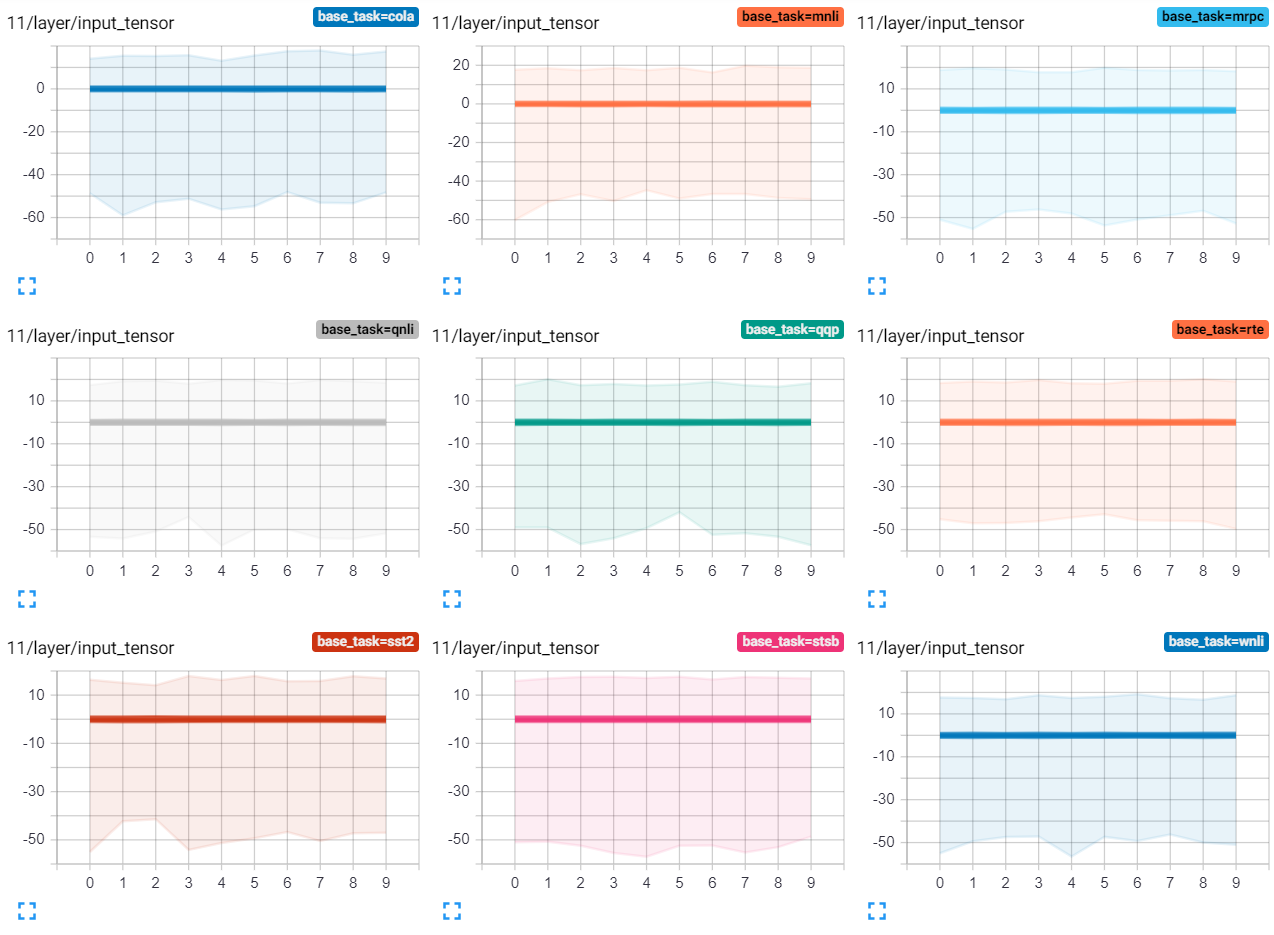}
    \label{fig:D_bert_base_input}
}
\hfill
\subfloat[output]{
    \includegraphics[width=.48\textwidth]{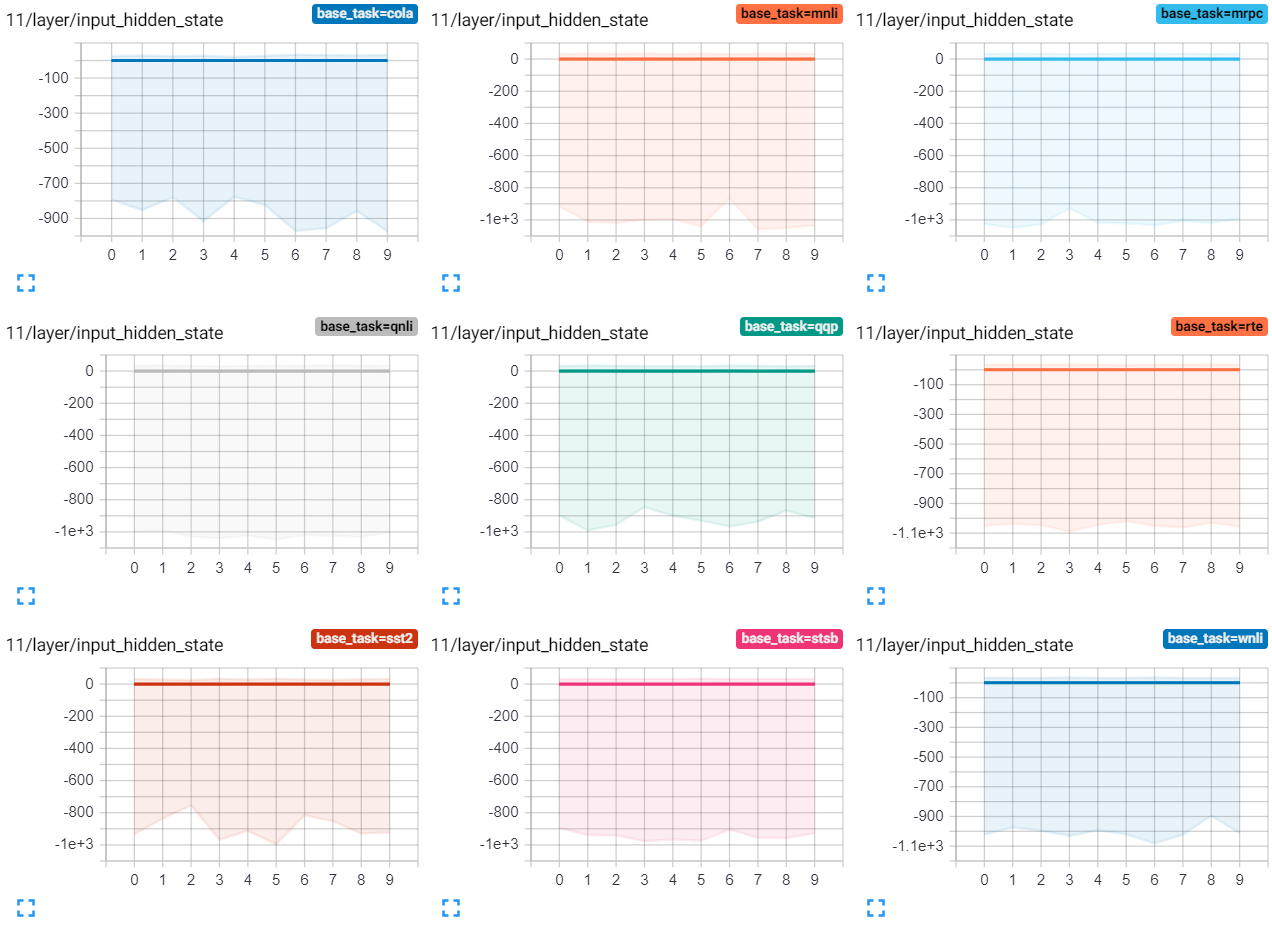}
    \label{fig:D_bert_base_output}
}
\caption{%
Activation distributions of FFN's input~\protect\subref{fig:D_bert_base_input} and output~\protect\subref{fig:D_bert_base_output} in second to the last layer for BERT-base, evaluated on first ten data sequences from development sets of GLUE downstream tasks (full-precision). 
In each sub-plot, left-to-right, top-to-bottom: CoLA, MNLI, MRPC $\rightarrow$ QNLI, QQP, RTE $\rightarrow$ SST-2, STS-B, WNLI.
\emph{x-axis}: index of data sequence. 
\emph{y-axis}: the range (note the scales are different for the input and the output).
}
\label{fig:D_bert_base}
\end{figure*}

\begin{figure*}[htb]
\subfloat[input]{
    \includegraphics[width=.48\textwidth]{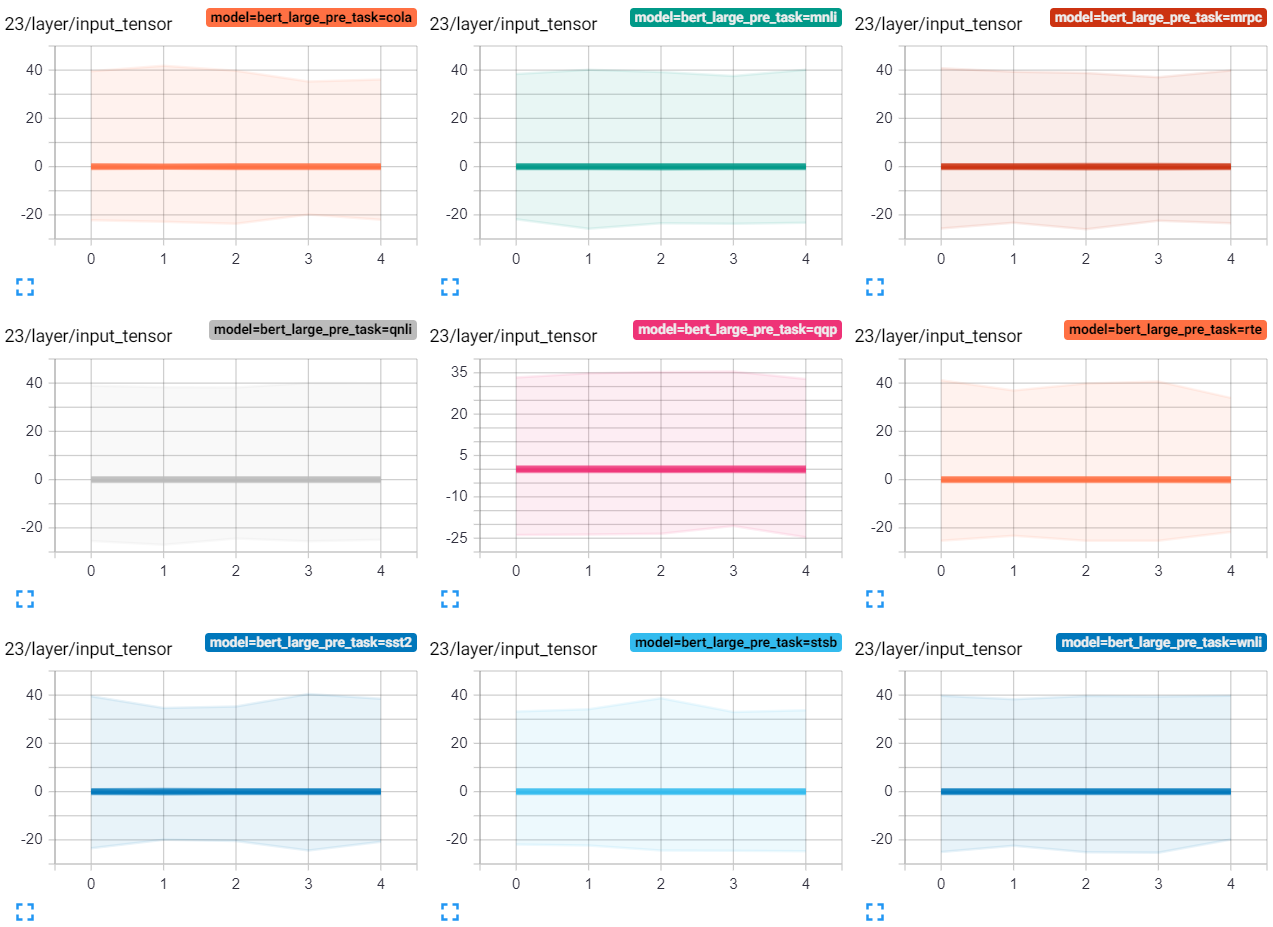}
    \label{fig:D_bert_large_input}
}
\hfill
\subfloat[output]{
    \includegraphics[width=.48\textwidth]{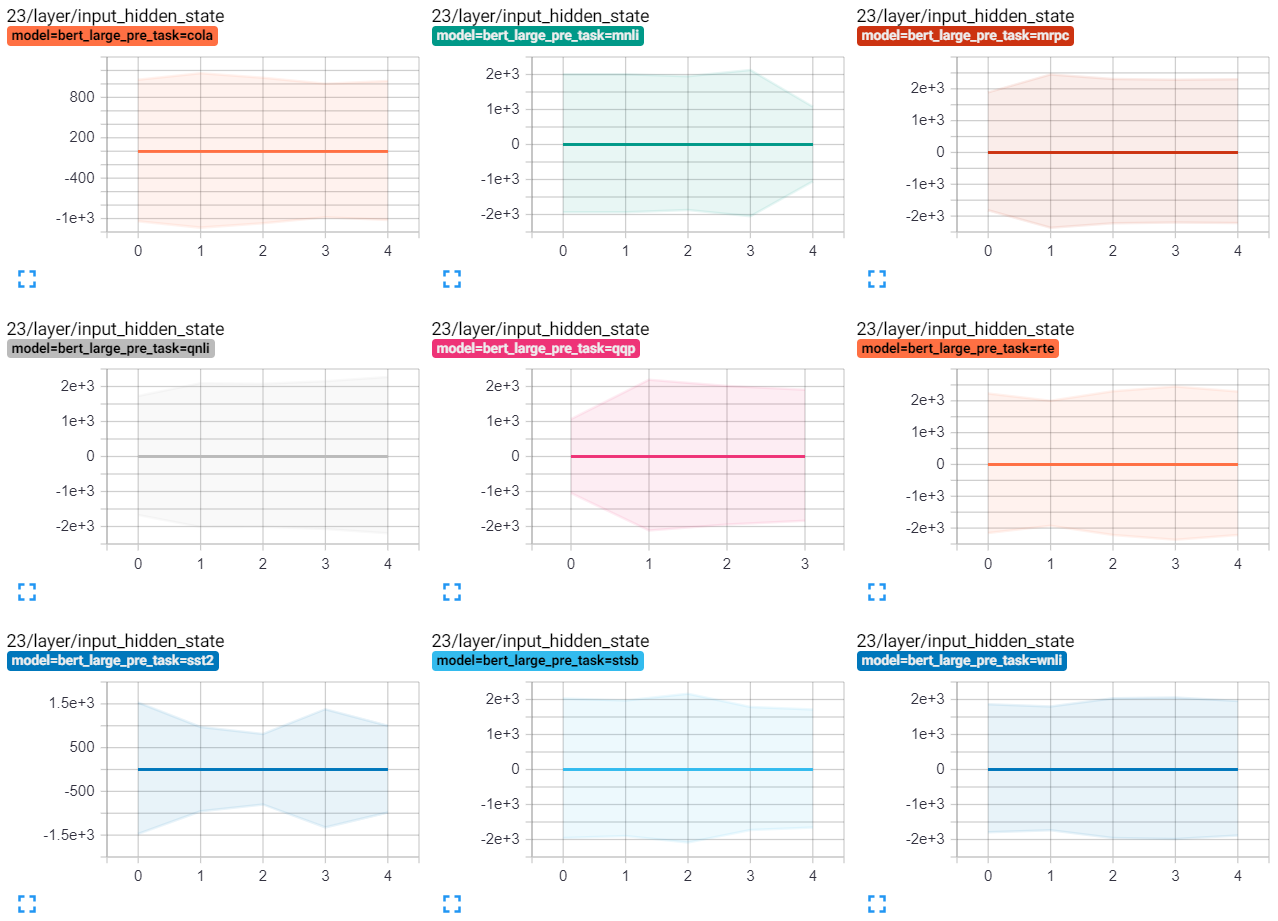}
    \label{fig:D_bert_large_output}
}
\caption{%
Activation distributions of FFN's input~\protect\subref{fig:D_bert_large_input} and output~\protect\subref{fig:D_bert_large_output} in second to the last layer for BERT-large, evaluated on first five data sequences from development sets of GLUE downstream tasks (full-precision). 
In each sub-plot, left-to-right, top-to-bottom: CoLA, MNLI, MRPC $\rightarrow$ QNLI, QQP, RTE $\rightarrow$ SST-2, STS-B, WNLI.
\emph{x-axis}: index of data sequence. 
\emph{y-axis}: the range (note the scales are different for the input and the output).
}
\label{fig:D_bert_large}
\end{figure*}

\begin{figure*}[htb]
\subfloat[input]{
    \includegraphics[width=.48\textwidth]{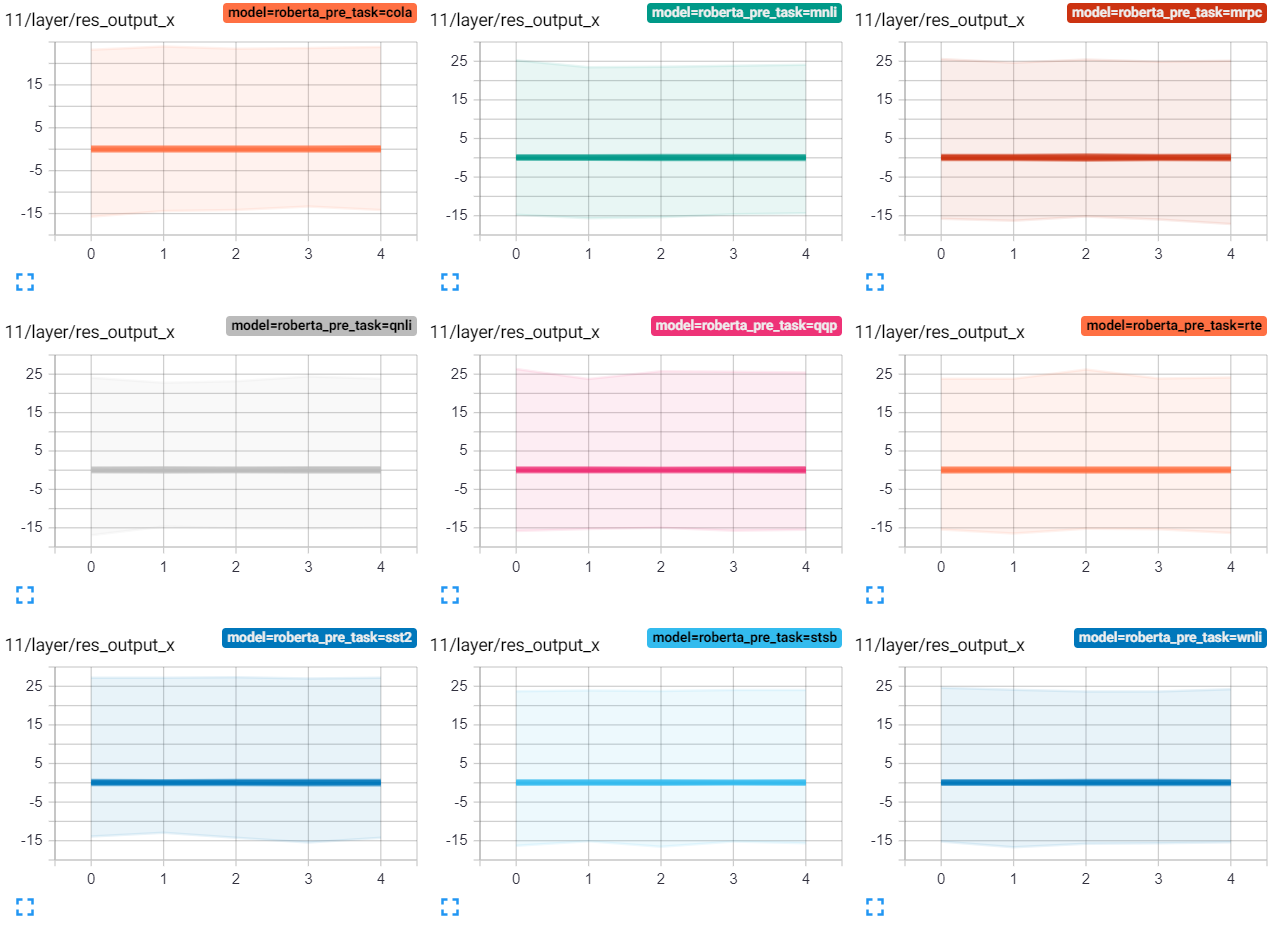}
    \label{fig:D_roberta_input}
}
\hfill
\subfloat[output]{
    \includegraphics[width=.48\textwidth]{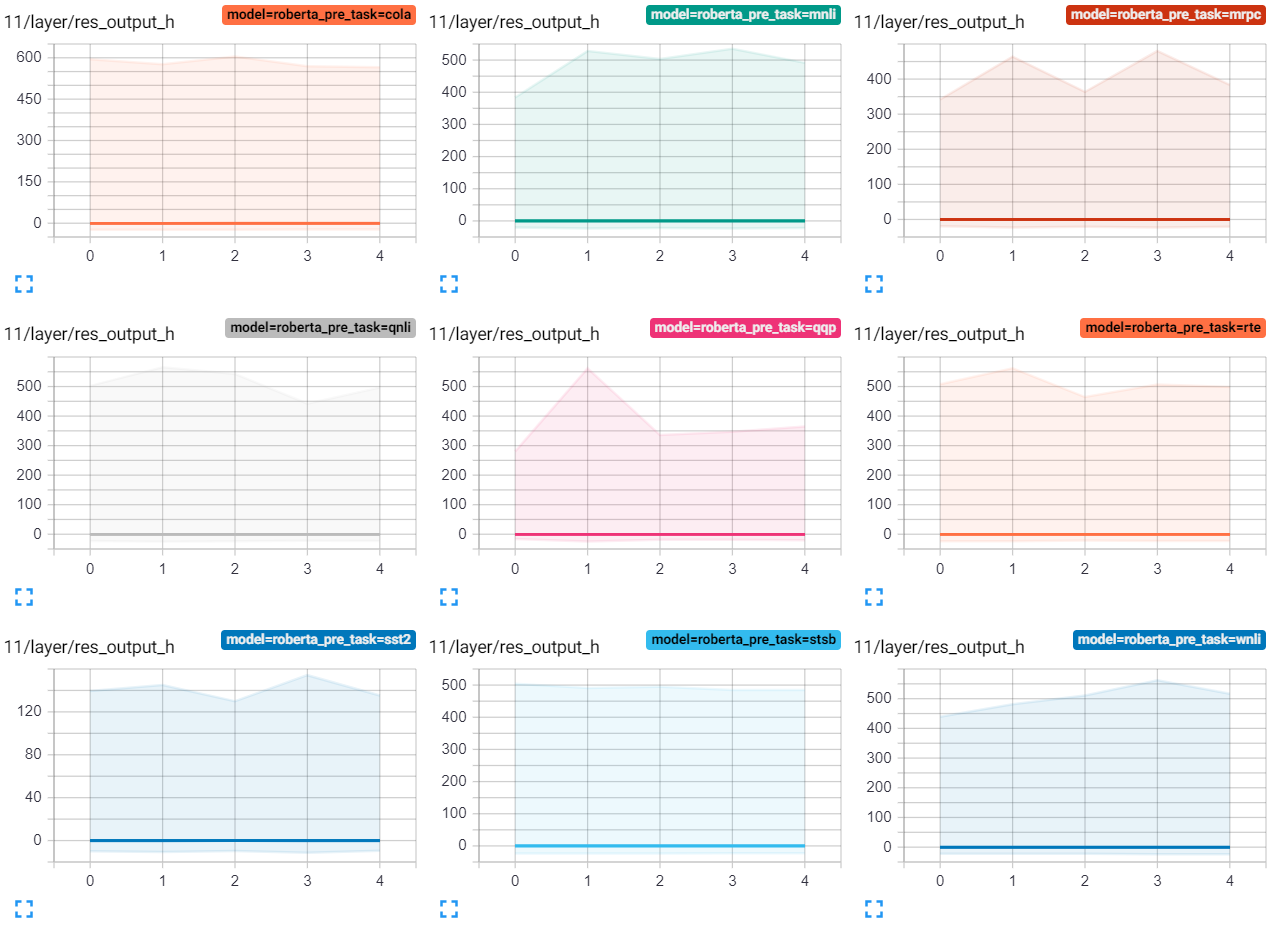}
    \label{fig:D_roberta_output}
}
\caption{%
Activation distributions of FFN's input~\protect\subref{fig:D_roberta_input} and output~\protect\subref{fig:D_roberta_output} in second to the last layer for RoBERTa-base, evaluated on first five data sequences from development sets of GLUE downstream tasks (full-precision). 
In each sub-plot, left-to-right, top-to-bottom: CoLA, MNLI, MRPC $\rightarrow$ QNLI, QQP, RTE $\rightarrow$ SST-2, STS-B, WNLI.
\emph{x-axis}: index of data sequence. 
\emph{y-axis}: the range (note the scales are different for the input and the output).
}
\label{fig:D_roberta}
\end{figure*}

\begin{figure*}[htb]
\subfloat[input]{
    \includegraphics[width=.48\textwidth]{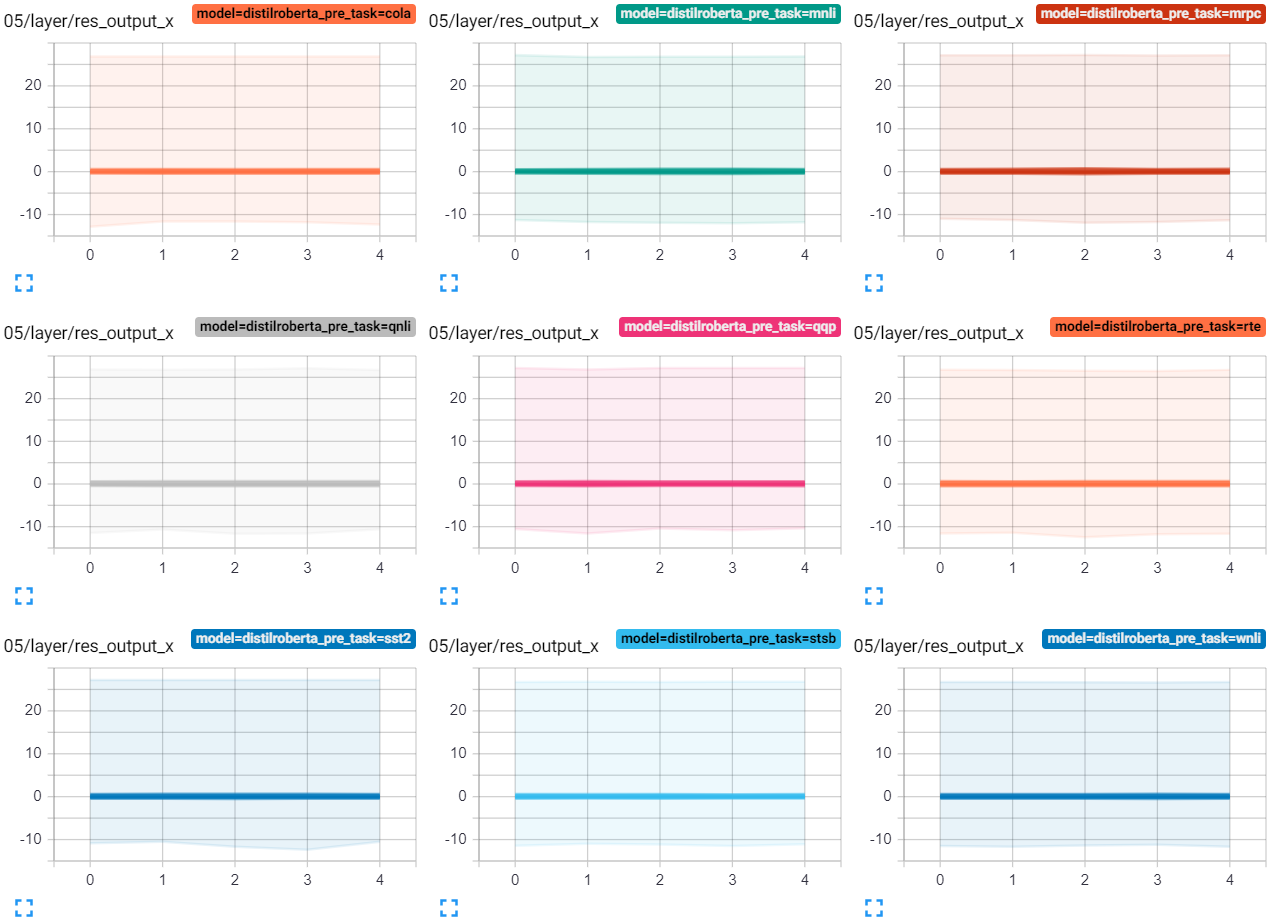}
    \label{fig:D_distilroberta_input}
}
\hfill
\subfloat[output]{
    \includegraphics[width=.48\textwidth]{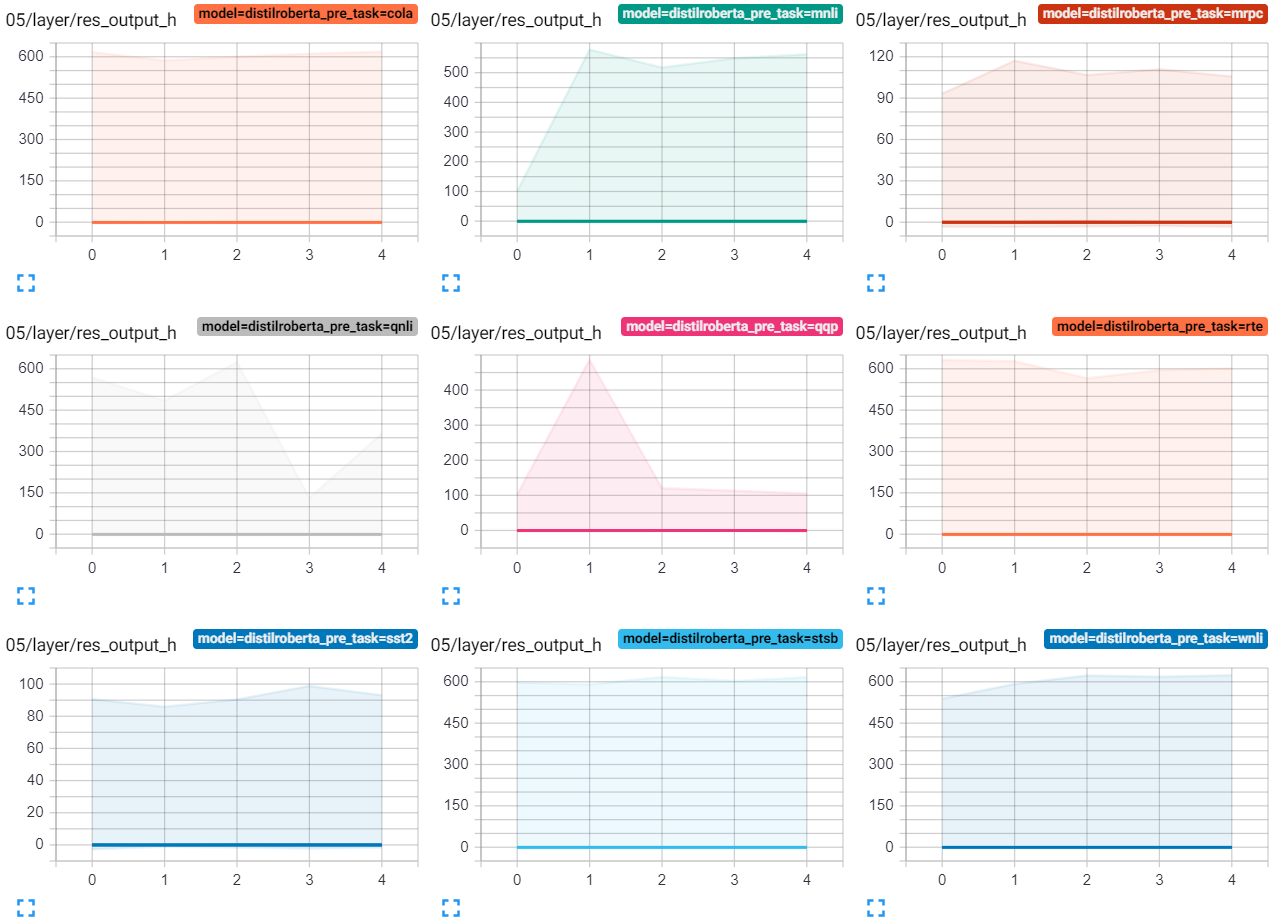}
    \label{fig:D_distilroberta_output}
}
\caption{%
Activation distributions of FFN's input~\protect\subref{fig:D_distilroberta_input} and output~\protect\subref{fig:D_distilroberta_output} in second to the last layer for DistilRoBERTa-base, evaluated on first five data sequences from development sets of GLUE downstream tasks (full-precision). 
In each sub-plot, left-to-right, top-to-bottom: CoLA, MNLI, MRPC $\rightarrow$ QNLI, QQP, RTE $\rightarrow$ SST-2, STS-B, WNLI.
\emph{x-axis}: index of data sequence. 
\emph{y-axis}: the range (note the scales are different for the input and the output).
}
\label{fig:D_distilroberta}
\end{figure*}

\begin{figure*}[htb]
\subfloat[input]{
    \includegraphics[width=.48\textwidth]{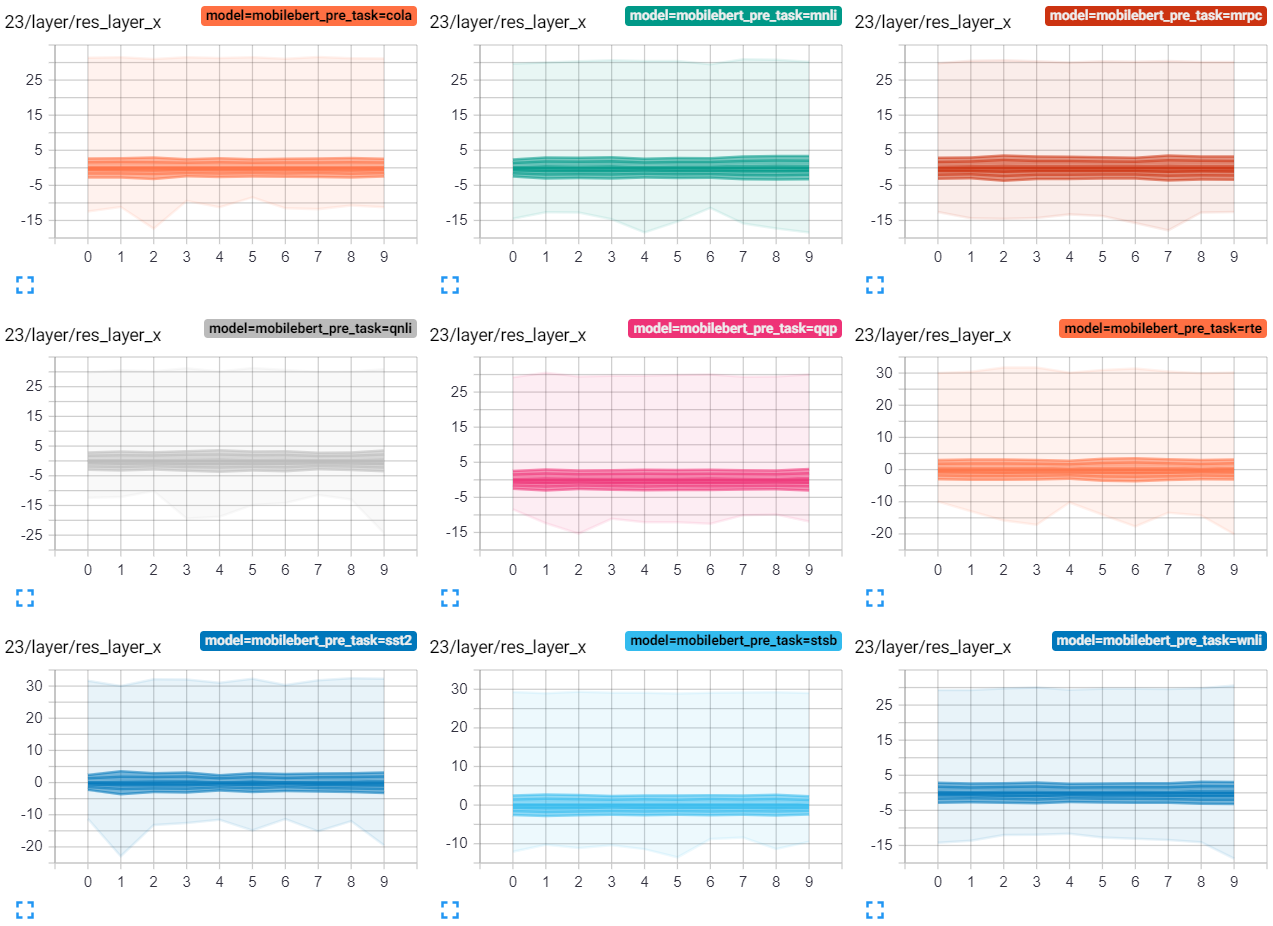}
    \label{fig:D_mobilebert_input}
}
\hfill
\subfloat[output]{
    \includegraphics[width=.48\textwidth]{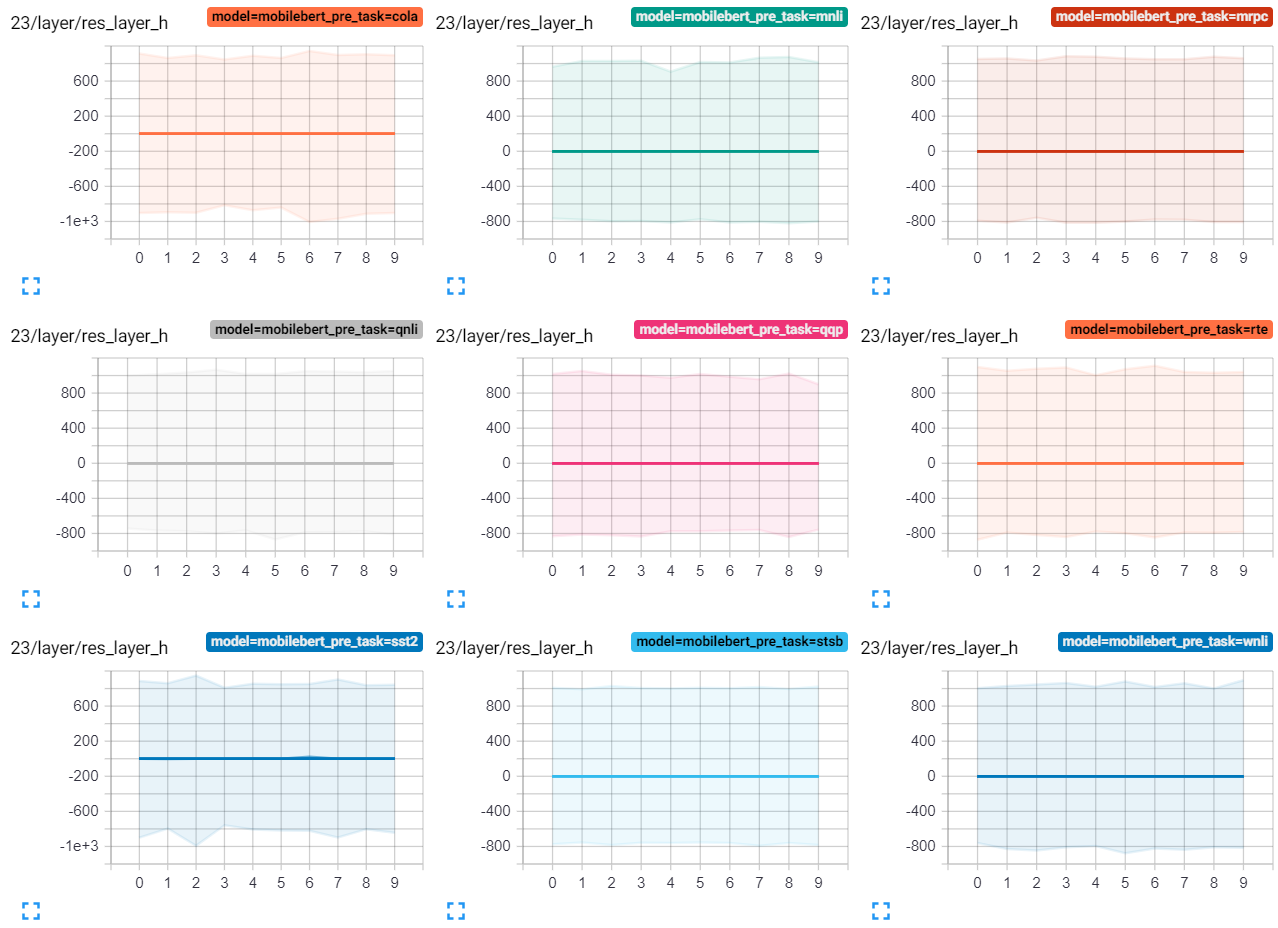}
    \label{fig:D_mobilebert_output}
}
\caption{%
Activation distributions of FFN's input~\protect\subref{fig:D_mobilebert_input} and output~\protect\subref{fig:D_mobilebert_output} in second to the last layer for MobileBERT-base, evaluated on first ten data sequences from development sets of GLUE downstream tasks (full-precision). 
In each sub-plot, left-to-right, top-to-bottom: CoLA, MNLI, MRPC $\rightarrow$ QNLI, QQP, RTE $\rightarrow$ SST-2, STS-B, WNLI.
\emph{x-axis}: index of data sequence. 
\emph{y-axis}: the range (note the scales are different for the input and the output).
}
\label{fig:D_mobilebert}
\end{figure*}

\begin{itemize}
    \item BERT-base in Figure~\ref{fig:D_bert_base},
    \item BERT-large in Figure~\ref{fig:D_bert_large},
    \item RoBERTa-base in Figure~\ref{fig:D_roberta},
    \item DistilRoBERTa-base in Figure~\ref{fig:D_distilroberta},
    \item MobileBERT-base in Figure~\ref{fig:D_mobilebert}.
\end{itemize}
In all cases, we used pre-trained checkpoints from HuggingFace library~\citep{wolf-etal-2020-transformers}.
%


\end{document}